\documentclass{ieeeaccess}

\usepackage{cite}
\usepackage{amsmath,amssymb,amsfonts}
\usepackage{graphicx}
\usepackage{textcomp}
\usepackage{mathtools}
\usepackage{physics}
\usepackage{booktabs}
\usepackage{multirow}
\usepackage{colortbl}
\usepackage{color}
\usepackage[font={sf,scriptsize},
            labelfont={bf,color=accessblue},
            caption=false]
           {subfig}

\def\BibTeX{{\rm B\kern-.05em{\sc i\kern-.025em b}\kern-.08em
    T\kern-.1667em\lower.7ex\hbox{E}\kern-.125emX}}

\newcommand{\defeq}{\vcentcolon=}

\DeclareMathOperator*{\argmax}{argmax}

\definecolor{lightred}{RGB}{243, 156, 169}
\definecolor{lightgreen}{RGB}{214, 235, 225}

\begin{document}

\history{}
\doi{}

\title{Quantum Deep Reinforcement Learning for Robot Navigation Tasks}
\author{%
    \uppercase{Hans Hohenfeld}\authorrefmark{1},
    \uppercase{Dirk Heimann}\authorrefmark{1},
    \uppercase{Felix Wiebe}\authorrefmark{2}, and
    \uppercase{Frank Kirchner}\authorrefmark{1,2}
}

\address[1]{%
    Robotics Research Group,
    University of Bremen,
    Robert-Hooke-Straße 1,
    28359 Bremen, Germany
    (e-mail: \{hans.hohenfeld, dirk.heimann, frank.kirchner\}@uni-bremen.de)
}
\address[2]{%
    Robotics Innovation Center (RIC),
    German Research Center for Artificial Intelligence (DFKI),
    Robert-Hooke-Straße 1,
    28359 Bremen, Germany
    (e-mail: \{felix.wiebe, frank.kirchner\}@dfki.de)
}

\tfootnote{%
This work was funded by the German Federal Ministry of Economic Affairs and Climate Action (BMWK)
and German Aerospace Center e.V. (DLR e.V.) through the project QINROS under project numbers
50RA2033 (DFKI) and 50RA2032 (University of Bremen).
}

\corresp{Corresponding author: Hans Hohenfeld (e-mail: hans.hohenfeld@uni-bremen.de).}

\begin{abstract}
    We utilize hybrid quantum deep reinforcement learning to learn navigation tasks for a simple,
    wheeled robot in simulated environments of increasing complexity. For this, we train
    parameterized quantum circuits (PQCs) with two different encoding strategies in a hybrid
    quantum-classical setup as well as a classical neural network baseline with the double deep Q
    network (DDQN) reinforcement learning algorithm. Quantum deep reinforcement learning (QDRL) has
    previously been studied in several relatively simple benchmark environments, mainly from the
    OpenAI gym suite. However, scaling behavior and applicability of QDRL to more demanding tasks
    closer to real-world problems e.\,g., from the robotics domain, have not been studied
    previously. Here, we show that quantum circuits in hybrid quantum-classic reinforcement learning
    setups are capable of learning optimal policies in multiple robotic navigation scenarios with
    notably fewer trainable parameters compared to a classical baseline. Across a large number of
    experimental configurations, we find that the employed quantum circuits outperform the classical
    neural network baselines when equating for the number of trainable parameters. Yet, the
    classical neural network consistently showed better results concerning training times and
    stability, with at least one order of magnitude of trainable parameters more than the
    best-performing quantum circuits. However, validating the robustness of the learning methods in
    a large and dynamic environment, we find that the classical baseline produces more stable and
    better performing policies overall. For the two encoding schemes, we observed better results for
    consecutively encoding the classical state vector on each qubit compared to encoding each
    component on a separate qubit. Our findings demonstrate that current hybrid quantum
    machine-learning approaches can be scaled to simple robotic problems while yielding sufficient
    results, at least in an idealized simulated setting, but there are yet open questions regarding
    the application to considerably more demanding tasks. We anticipate that our work will
    contribute to introducing quantum machine learning in general and quantum deep reinforcement
    learning in particular to more demanding problem domains and emphasize the importance of
    encoding techniques for classic data in hybrid quantum-classical settings.
\end{abstract}


\titlepgskip=-21pt
\maketitle

\section{Introduction}%
\label{sec:intro}

    \begin{figure}[t]
        \centering
        \includegraphics{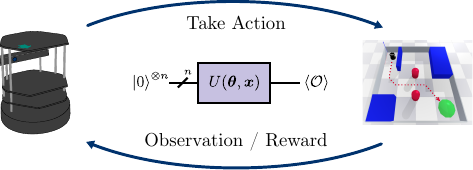}
        \caption{Main contribution: We use parameterized quantum circuits (PQCs) as
        function approximators in the DDQN Deep Reinforcement Learning algorithm to learn optimal
        policies for a simulated Turtlebot robotic system in several simulated navigation tasks.}
        \label{fig:contrib}
    \end{figure}

    Robotics research and applications pose various algorithmic challenges, ranging from large-scale
    optimization, processing of high-dimensional sensory input, planning the execution of complex
    tasks in demanding environments, and learning of autonomous, adaptable behaviors. On the latter,
    deep reinforcement learning is used to produce impressive results in tasks such as learning
    complex manipulation behaviors~\cite{gu2017}, reaching, tracking and,
    navigation~\cite{mahmood2018}, manipulation based on visual input~\cite{kalashnikov2018} as well
    as dexterous hand movements~\cite{andrychowicz2020} among many others. It constitutes a central
    role on the path toward autonomous and life-long learning robots.

    Quantum computing algorithms~\cite{montanaro2016} present a novel way of approaching algorithmic
    problems and offer theoretical advantages over classical algorithms for specific problems like
    factoring numbers~\cite{shor1994}, unstructured search~\cite{grover1996}, and solving systems of
    linear equations~\cite{harrow2009}. With more development and further resources, quantum
    computing and, in particular quantum machine learning~\cite{alchieri2021} may contribute to the
    development of artificial intelligence in general and the learning of autonomous behaviors for
    robots in particular~\cite{kirchner2020}.

    The idea of robots controlled by quantum computers, interacting with an environment on the scale
    of individual quantum states has arguably first been hypothesized and described by quantum
    computing pioneer Paul Benioff in the late 1990s and early 2000s~\cite{benioff1997,
    benioff1998, benioff2000, benioff2002}. While those envisioned \textit{Quantum Robots} are very
    different from typical mechanical robotic systems as they can be found in various practical
    applications today, the idea of a mobile system utilizing quantum computing hardware remains
    intriguing.

    Quantum computing technology has not yet reached the state of mobile, embedded, and potentially
    battery-powered quantum hardware but has made remarkable progress over the last two decades.
    Research institutions and companies are building quantum computers with increasing capabilities,
    and while current Noisy Intermediate-Scale Quantum Computers (NISQ) are limited in the number of
    qubits, coherence times, and fidelity of operation~\cite{preskill2018}, they already enable
    exploring solutions for various problems~\cite{brooks2019}.

    One potential application for NISQ devices is the hybrid training of parameterized quantum
    circuits (PQCs) as machine learning models~\cite{farhi2018}. While this technique has been
    studied in various domains of machine learning~\cite{benedetti2019}, deep reinforcement learning
    has only recently attracted substantial research interest in this context. Existing works~(see
    Sec.~\ref{sec:sota_qdrl}) demonstrate the applicability of hybrid quantum-classical approaches
    for reinforcement learning tasks, with performances similar to classical algorithms while
    learning notably more compact models. However, their scope is currently limited to relatively
    simple benchmark environments, mainly from the OpenAI gym suite~\cite{brockman2016}.

    Or main contributions, illustrated in Fig.~\ref{fig:contrib}, are as follows. We demonstrate the
    feasibility of quantum deep reinforcement learning in three simulated robotic navigation tasks
    of increasing size and difficulty. Thereby, we extend the scope of previously introduced methods
    to substantially more complex tasks in the robotic domain, as we show by comparative experiments
    with typical benchmark environments. Furthermore, we compare to different encoding strategies
    for the classical state of the robot into a quantum circuit and also analyze the scaling
    behavior of the quantum circuits relative to a classical baseline. To validate the robustness of
    the presented methods, we additionally demonstrate their application in a substantially larger,
    more demanding and dynamic environment. In comparison to previous works in the field of quantum
    deep reinforcement learning, we thereby increase the complexity of considered learning tasks and
    furthermore provide a systematic evaluation of the scaling behaviour of quantum models in this
    context.  Finally, we discuss various challenges and limitations of quantum deep reinforcement
    learning in a robotic context, as well as potential areas of research for quantum machine
    learning to contribute to the future advancements in autonomous robotics.

    The rest of this paper is outlined as follows: In Sec.~\ref{sec:related}, we provide an
    overview of previous works regarding deep reinforcement learning with PQCs. Subsequently, we
    outline the quantum deep reinforcement learning framework underlying this work in
    Sec.~\ref{sec:qdrl}. Afterwards, the learning setup with regard to the simulated
    environments and learning methods is documented in Sec.~\ref{sec:method}. We present the
    training results of the suggested methods compared to a classical baseline in
    Sec.~\ref{sec:results} before summarizing our main findings and discussing their implications
    and limitations in Sec.~\ref{sec:discussion}. Finally, we give an outlook toward potential
    future research directions in Sec.~\ref{sec:outlook}.

\section{Related Work}
\label{sec:related}
    Introducing quantum algorithmic techniques and quantum mechanical effects into reinforcement
    learning (RL) methods is an active and growing field of research. Meyer \textit{et
    al.}~\cite{meyer2022} give an overview over various proposed methods and applications in this
    area. In the following, we highlight important methods and results from this line of research.

    \subsection{Quantum RL and Quantum inspired RL}
    Quantum mechanics and quantum computing were introduced reinforcement learning by
    Dong~\textit{et al.}~\cite{dong2008}, who proposed Quantum Reinforcement Learning (QRL). In the
    QRL algorithm, the classical states and actions of the agent are expressed in the orthonormal
    eigenbasis of a Hermitian observable. Actions are chosen by measuring in that basis from a
    superposition state, where the amplitudes of that superposition state are modified during
    learning utilizing amplitude amplification, the essential building block of Grover's
    algorithm~\cite{grover1996}. The authors evaluate the QRL algorithm in a discrete maze world,
    comparing it to the tabular TD(0) RL algorithm~\cite{sutton2018}, achieving convincing
    performance. Quantum-inspired Reinforcement Learning (QiRL)~\cite{dong2012} is a classical RL
    algorithm that builds on the ideas of QRL, using a quantum-inspired probabilistic sampling
    technique to address the exploration vs.\ exploitation~\cite{sutton2018} problem in RL and a
    classical technique inspired by amplitude amplification to control the sampling probabilities.
    The algorithm is demonstrated on a simulated grid world and real-world robot navigation task
    with a wheeled MT-R robot. A variant of QiRL with flexible rotation angles in the amplitude
    amplification step is proposed in Ref.~\cite{li2021}, which shows better performance on a UAV
    navigation problem compared to tabular Q-learning~\cite{watkins1989} with two different
    exploration strategies. Hu~\textit{et al.}~\cite{hu2021} apply QRL to the \texttt{MountainCar}
    and \texttt{CartPole} problems from the OpenAI Gym~\cite{brockman2016} suite, focusing on the
    exploration vs.\ exploitation problem, finding better overall learning performance compared to
    the classical Q-learning algorithm with an $\epsilon$-greedy policy. Quantum-inspired Experience
    Replay (QER)~\cite{wei2022} is an extension of the concepts of QiRL to the representation of
    experiences and sampling from the replay buffer in Deep Reinforcement Learning (DRL), which the
    authors evaluate in several \texttt{Atari 2600} game environments~\cite{brockman2016} and
    compare to baseline experience replay and prioritized experience replay with several variants of
    the DQN~\cite{hu2021} algorithm.

    This line of work with QRL and QiRL emphasizes the expression of states and actions in RL
    problems in quantum states, efficiently updating measurement probabilities leveraging amplitude
    amplification, and expressing the same concepts in a classical learning setup. QER extends
    these ideas to experience replay in DRL. Our contribution is conceptually different, as we
    focus on substituting classical neural networks in DRL with parameterized quantum circuits
    while keeping the learning algorithm and representation of all aspects of the learning task
    unchanged.

    \subsection{Quantum Environments}
    Dunjko \textit{et al.}~\cite{Dunjko2016} propose a quantum-enhanced framework that, in
    principle, covers supervised, unsupervised, and reinforcement learning but, in its formulation,
    is closest to the latter. In this framework, the agent and the environment exchange actions and
    percepts by applying completely positive trace-preserving (CPTP) maps to a shared quantum
    register and their local quantum memory. The authors analyze the conditions under which an agent
    in this framework can outperform its classical counterpart. They extend this work to a
    meta-learning scenario in Ref.~\cite{Dunjko2017} demonstrating further improvements.

    Saggio \textit{et al.}~\cite{Saggio2021} suggest a reinforcement learning setting in which the
    agent and environment exchange information on a classical and a quantum channel in an
    alternating way and show how, in such a setting, an agent performs better than with strictly
    classical information exchange. The authors validate the concept by performing experiments on a
    photonic quantum processor. Theoretical performance analysis of this framework is provided in
    Ref.~\cite{Hamann2022}, where the authors find a quadratic learning speed-up, which still holds
    under hardware noise and limited coherence times. Dalla \textit{et al.}~\cite{dalla2022}
    successfully demonstrate a classical deep reinforcement learner in a quantized maze environment
    based on quantum walks, considering potentially noisy dynamics.

    Quantum-enhanced learning frameworks, as considered in these works, require some form of quantum
    information based interaction between agent and environment and are, in that aspect, different
    from our contribution. We consider an environment from the robotic domain, where
    agent-environment interaction is strictly classical.

    \subsection{Projective Simulation}
    Projective simulation~\cite{briegel2012} is an extension of the RL learning framework by an
    episodic and compositional memory, which allows the agent to predict potential future events
    using random walks on that memory. In Ref.~\cite{paparo2014}, the authors propose an extension of
    this learning method using quantum walk on quantum memory instead to achieve a quadratic
    speed-up, which was later demonstrated in a proof-of-principle experiment on an ion-trap based
    quantum system by Sriarunothai \textit{et al.}~\cite{Sriarunothai2018}.

    The deep reinforcement learning algorithm we use in our work does not utilize any form of
    episodic memory, hence the suggested techniques in this line of research are not immediately
    applicable.

    \subsection{Quantum Deep Reinforcement Learning}
    \label{sec:sota_qdrl}
    In quantum deep reinforcement learning (QDRL), the line of research from which our contribution
    originates, one or multiple classical neural networks are replaced or extended by parameterized
    quantum circuits. In contrast, agent-environment interaction and as the learning procedure are
    kept classical. We give a detailed account of the underlying theory in Sec.~\ref{sec:qdrl}. The
    focus in this relatively new field so far has mostly been on showing the feasibility of the
    methods, understanding their capabilities and limitations, as well as finding quantum-classical
    separations in learning tasks.

    In several works the Q-function approximation in the DQN algorithm is implemented by a PQC.
    Chen \textit{et al.}~\cite{chen2020} use basis encoding~\cite{schuld2021} followed by CNOT
    entanglements and parameterized Pauli rotations without a data re-uploading structure for the
    \texttt{FrozenLake}~\cite{brockman2016} and a \texttt{CognitiveRadio} task~\cite{gawlowicz2019}
    with discrete state and action spaces. Lockwood \textit{et al.}~\cite{lockwood2020} use a
    different encoding technique and combine the parameterized circuit with quantum pooling
    operations~\cite{cong2019} and classical neural network layers without data re-upload. This
    setup is able to learn a \texttt{Blackjack} environment but do not successful learn
    \texttt{Cartpole-v0}~\cite{brockman2016}. In Ref.~\cite{skolik2021} the circuit layout and
    encoding scheme is similar to the one that we employ in this work. The architecture also
    includes data re-uploading and enables learning on \texttt{FrozenLake} and \texttt{Cartpole-v0}.

    In addition, PQCs have also been used in the policy gradient methods
    REINFORCE~\cite{sutton2018}. In Ref.~\cite{jerbi2021} the PQC architecture also includes data
    re-upload scheme. Included in the REINFORCE algorithm, the setup is able to solve
    \texttt{Cartpole-v1}, \texttt{Mountaincar-v0} and \texttt{Acrobot-v1}. Additionally, the authors
    demonstrate experimentally and formally that hybrid quantum deep reinforcement learning can
    solve environments based on the discrete logarithm problem~\cite{liu2021} which are intractable
    for classical learning methods. A variant of the REINFORCE algorithm is used in
    Ref.~\cite{sanches2022} to optimize PQCs, which replace the classical attention head layers
    originally introduced in Ref.~\cite{kool2018}, to solve the vehicle routing problem and achieve
    similar results as the classical counterpart.

    Furthermore, actor critic methods such as proximal policy optimization (PPO)~\cite{schulman2017}
    and soft actor-critic (SAC)~\cite{haarnoja2018} have also been adapted with PQCs. In
    Ref.~\cite{kwak2021} the PPO algorithm is augmented with PQCs by exchanging the actor
    approximation network.  The PQC has no data re-uploading scheme and is trained on
    \texttt{Cartpole-v0} without completely solving it. In Ref.~\cite{hsiao2022} unentangled PQCs
    with a fully connected classical layer as post processing unit replace the classical estimator
    for the actor and critic. This setup solves OpenAI Gym environments \texttt{Cartpole-v1},
    \texttt{Acrobot-v1} and \texttt{LunarLander-v2}.

    Nagy \textit{et al.}~\cite{Nagy2021} simulate a hybrid quantum version of PPO on a photonic
    processor which demonstrates that PQC equivalences on photonic quantum computers can be used for
    reinforcement learning as well. In Ref.~\cite{lan2021} the author demonstrates that the critic
    network in SAC can be exchanged with a PQC followed by a classical neural network and still
    solve the \texttt{Pendulum-v0} problem from OpenAI gym with continuous state and action spaces.

    Several works introduce parameterized quantum circuits into a hybrid quantum-classical learning
    setup, without strictly falling into the category of deep reinforcement learning.  Cherrat
    \textit{et al.}~\cite{Cherrat2022} implement a quantum version of policy iteration to solve
    \texttt{FrozenLake} and the \texttt{InvertedPendulum} environment. Franken \textit{et
    al.}~\cite{franken2022} implement a gradient-free method based on evolutionary methods to
    optimize a PQC that receives input data encoded by a tensor network. This setup is able to solve
    \texttt{MiniGrid} worlds~\cite{gym_minigrid} with discrete state space.

    Or contribution extends upon these previous works in the following way:
    \begin{itemize}
        \item We extend the scope of QDRL to considerably more complex learning tasks from the
            robotic domain. We establish that increase in complexity by comparative experiments (see
            Appendix~\ref{sec:app_baseline}).
        \item We systematically evaluate the scaling behaviour of parameterized quantum circuits in
            QDRl across task complexity as well as model size, which has previously not been done.
        \item We compare different encoding strategies suggested in the literature for re-uploading
            circuits with regards to their performance in a QDRL scenario.
    \end{itemize}
    Thereby we extend the understanding of the feasibility of QDRL from very simple benchmark
    environments towards more realistic application scenarios from the robotics domain and
    contribute to the understanding of the model scaling behaviour in this context.

\section{Quantum Deep Reinforcement Learning}%
\label{sec:qdrl}%

    \subsection{Double Deep Q-Networks}
    \begin{figure}[t]
        \centering
        \includegraphics{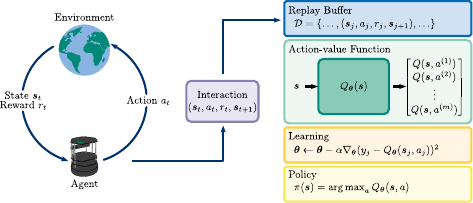}
        \caption{Reinforcement Learning setup (left) and main parts of the DDQN algorithm
            (right). An agent interacts with an environment by performing action $a_{t}$ after
            observing a state of the environment $\boldsymbol{s}_{t}$, causing a transition to state
            $\boldsymbol{s}_{t+1}$ and receiving a reward $r_{t}$. For the DDQN algorithm, these
            interactions are stored in a replay buffer, from which regularly random mini-batches are
            sampled to train an artificial neural network $Q_{\boldsymbol{\theta}}$ approximating
            the action-value function.
        }
        \label{fig:rl}
    \end{figure}

    For all our experiments, we used the Double Deep Q-Network~(DDQN)~\cite{hasselt2016} algorithm
    as it performed slightly better on average compared to e.\,g., the basic Deep Q-Network
    algorithm (DQN)~\cite{mnih2015}. DDQN is a model-free, off-policy deep RL algorithm that uses a
    neural network to approximate the Q-function from the basic Q-learning
    algorithm~\cite{watkins1989}.

    RL is used to solve Markov Decision Processes (MDPs), that is, discrete-time,
    stochastic processes $(S, A, T, r, p_{0})$ with
    \begin{itemize}
        \item $S$: The state space, a set of all possible states of an environment
        \item $A$: The action space, a set of all possible actions for an agent
        \item $T:S\times A\times S \to [0,1]$: The possibly stochastic transition function with
            $T(\boldsymbol{s},a,\boldsymbol{s}^{\prime})=p(\boldsymbol{s}^{\prime}|\boldsymbol{s}, a)$
            being the probability of transitioning to state $\boldsymbol{s}^{\prime}$ after taking
            action $a$ in state $\boldsymbol{s}$.
        \item $r:S\times A\times S\to \mathbb{R}$: A reward function with
            $r(\boldsymbol{s},a,\boldsymbol{s}^{\prime})$ denoting a
            numeric reward for taking action $a$ in state $\boldsymbol{s}$ and transitioning to
            state $\boldsymbol{s}^{\prime}$ and
        \item $p_{0}$: A probability for each state to be a starting state of the MDP.
    \end{itemize}

    The general scheme of interaction for an agent in an environment governed by an MDP is
    illustrated on the left side of Fig.~\ref{fig:rl}. At each time step $t$, the agent observes
    a state $\boldsymbol{s}_{t}$ of the environment, takes an action $a_{t}$, which causes
    a transition to state $\boldsymbol{s}_{t+1}$ and the agent to receive a reward $r_{t}$.
    The agent's action selection is governed by a policy $\pi:S\to A$ and the goal is to maximize
    the total cumulative reward $R$ for a possibly infinite time horizon, given by
    \begin{equation}
        R = \sum_{t=0}^{\infty}\gamma^{t}r_{t}
    \end{equation}
    with $\gamma\in[0,1]$, called discount factor, encoding the preference for immediate over
    long-term rewards.

    In the DDQN algorithm, this is achieved by learning an optimal action-value function $Q:S\times
    A\to\mathbb{R}$. The action-value function $Q(\boldsymbol{s}, a)$ expresses the expected total
    cumulative reward for taking action $a$ in state $\boldsymbol{s}$
    \begin{equation}
        Q(\boldsymbol{s}, a) \defeq \langle R\rangle_{\boldsymbol{s},a,\pi},
    \end{equation}
    and the greedy policy can be expressed in terms of $Q(\boldsymbol{s}, a)$ by
    \begin{equation}
        \label{eq:greedy}
        \pi(\boldsymbol{s}) \defeq \argmax_{a}Q(\boldsymbol{s}, a).
    \end{equation}

    The action-value function, also referred to as Q-function, is approximated by an artificial
    neural network $Q_{\boldsymbol{\theta}}$ with parameters $\boldsymbol{\theta}$. The neural
    network takes a state $\boldsymbol{s}\in S$ as input and computes $Q(\boldsymbol{s}, a^{(i)})$
    for all $a^{(i)}\in A$ as output. During learning, an $\epsilon$-greedy policy is employed by
    the agent, that is at each time step $t$ with probability $\epsilon\in[0,1]$, the agent takes
    a random action from $A$ to further explore the environment and with probability $1-\epsilon$, it
    follows the greedy policy \eqref{eq:greedy} to exploit its current knowledge. At the beginning
    of training, $\epsilon$ is commonly chosen with a value close to $1$ and gradually reduced
    toward $0$ as learning progresses.

    Interactions $(\boldsymbol{s}_t, a_t, r_{t+1}, \boldsymbol{s}_{t+1})$ are stored in a replay
    buffer~\cite{lin1992} from which at a predefined interval e.\,g., at each time step, a
    mini-batch is sampled to update $Q_{\boldsymbol{\theta}}$ with stochastic gradient descent,
    minimizing the loss
    \begin{equation}
        \mathcal{L}(\boldsymbol{\theta}) = (y_t - Q_\theta(\boldsymbol{s}_t, a_t))^2,
    \end{equation}
    with $y_t$ given by
    \begin{equation}\label{eq:ddqn}
        y_t = r_{t+1} + \gamma Q_{{\boldsymbol{\theta}}^\prime}( \boldsymbol{s}_{t+1},
            \argmax_{a^\prime} Q_{{\boldsymbol{\theta}}} (\boldsymbol{s}_{t+1}, a^\prime ) ).
    \end{equation}
    The target network $Q_{{\boldsymbol{\theta}}^\prime}$ is used to stabilize the training
    process~\cite{mnih2015}. It has the identical structure as $Q_{\boldsymbol{\theta}}$ and is
    periodically updated with $\boldsymbol{\theta}^\prime \leftarrow \boldsymbol{\theta}$.

    \subsection{Quantum Computing}
    We give a short introduction to the common notation of quantum computing and refer the interested
    reader to~\cite{nielsen2010} for a comprehensive explanation of basic and advanced concepts of
    this topic. The fundamental objects in quantum computing are qubits, analogous to
    bits in classical computing. Unlike classical bits, which can be in one of the two states, $0$ and
    $1$, a qubit can be in a state, which is a linear combination of those states. Using the bra-ket
    notation, a qubit state $\ket{\Psi}$ can be written as
    \begin{equation}
        \ket{\Psi} = \alpha \ket{0} + \beta \ket{1},
            \text{with } \alpha, \beta \in \mathbb{C} \text{, } |\alpha|^2 + |\beta|^2 = 1,
    \end{equation}
    where $\ket{0}$ and $\ket{1}$ are basis states of the underlying single-qubit Hilbert Space.
    During the probabilistic measurement process, the qubit will collapse to one of the two basis
    states, and $|\alpha|^2$ and $|\beta|^2$ can be interpreted as the probabilities for the
    respective basis states. Before the measurement, the state can be modified by applying quantum
    gates $U$
    \begin{equation}
        \ket{\Psi'} = U \ket{\Psi},
    \end{equation}
    formally described by unitary operators $U$. This formulation can be extended to
    multi-qubit systems by preparing an $n$-qubit quantum register. For quantum computers, this is
    commonly initialized in its computational basis state $\ket{0}^{\otimes n}$.

    \subsection{Parameterized Quantum Circuits for Deep Reinforcement Learning}

    \begin{figure}[t]
        \centering
        \includegraphics{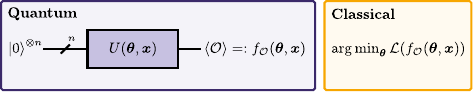}
        \caption{Basic principle of a parameterized quantum circuit as function approximator.
            A unitary $U(\boldsymbol{\theta}, \boldsymbol{x})$, which may be composed of
            any number of quantum gates, is applied to an $n$ qubit register initialized in its
            basis state. The unitary is parameterized by trainable parameters $\boldsymbol{\theta}$
            and input data $\boldsymbol{x}$. Thereby, the expectation value of an observable
            $\langle \mathcal{O}\rangle$ can be defined as a parameterized function
            $f_{\mathcal{O}}(\boldsymbol{\theta}, \boldsymbol{x})$. The parameters
            $\boldsymbol{\theta}$ are optimized toward a desired outcome, by minimizing a
            task specific loss $\mathcal{L}$ using a classical optimization technique e.\,g., gradient
            descent.}
        \label{fig:pqc}
    \end{figure}

    Variational quantum algorithms are a promising method to implement algorithms on current and
    near-term quantum computers as they are well suited for systems with a relatively small number of
    qubits, noisy operations, and limited coherence times~\cite{cerezo2021vqa}. Their basic principle
    of operation is the combination of a parameterized quantum circuit whose parameters are adjusted
    by a classical optimizer toward the desired outcome while evaluating the quantum circuit with
    adjusted parameters at each optimization step~\cite{mcclean2016}. First introduced in the
    context of variational quantum eigensolvers~\cite{peruzzo2014}, they became a major research
    area in quantum machine learning~\cite{cerezo2021vqa}.

    A parameterized quantum circuit (PQC) is a series of unitary quantum gates $U(\boldsymbol{\theta},
    \boldsymbol{x})$, which is applied to the computational basis of $n$ qubits $\ket{0}^{\otimes
    n}$. These gates are parameterized by variational parameters $\boldsymbol{\theta}$ and
    classical input data $\boldsymbol{x}$. Fig.~\ref{fig:pqc} shows this general ansatz for a
    machine learning application.

    The PQC's quantum state $\ket{U(\boldsymbol{\theta}, \boldsymbol{x})} = U(\boldsymbol{\theta},
    \boldsymbol{x})\ket{0}^{\otimes n}$ is computed and measured for many repeated iterations to
    gather sufficient statistics for the expectation value
    \begin{equation}
        \langle\mathcal{O}\rangle =
        \bra{U(\boldsymbol{x}, \boldsymbol{\theta})} \ \mathcal{O} \ \ket{U(\boldsymbol{\theta},
        \boldsymbol{x})}
    \end{equation}
    for an observable $\mathcal{O}$. $\bra{U(\boldsymbol{x},
    \boldsymbol{\theta})}$ denotes the conjugate transpose of $\ket{U(\boldsymbol{\theta},
    \boldsymbol{x})}$.
    The measured expectation value of the quantum computation can be interpreted as the
    computation of a parameterized function $f_{\mathcal{O}}({\boldsymbol{\theta}}, \boldsymbol{x})$
    depending on the observable, circuit parameters, and input.

    The parameters $\boldsymbol{\theta}$ are tuned with an appropriate method to fit a target
    function. E.g., in the domain of supervised machine learning, a loss function $\mathcal{L}$ can
    be minimized by performing gradient descent. Several analytic and numeric methods enable
    calculating gradients of quantum circuits with respect to their parameters, such as the finite
    difference methid~\cite{khan1999}, the parameter shift rule~\cite{mitarai2018} and adjoint
    differentiation~\cite{jones2020}.

    Various encoding methods for quantum machine learning tasks have been
    suggested~\cite{schuld2021}, but recent results on the expressiveness of quantum circuits
    emphasize the advantages of repeated encodings, also referred to as data
    re-upload~\cite{perez-salinas2020, schuld2021a}. Such an ansatz enables the circuit to compute
    functions of the form
    \begin{equation}
        f(\boldsymbol{\theta}, \boldsymbol{x}) =
            \sum_{\omega\in\Omega} c_\omega(\boldsymbol{\theta})e^{i\omega \boldsymbol{x}},
    \end{equation}
    which is a partial Fourier series with frequency spectrum $\Omega$ depending on the data encoding
    and coefficients $c_\omega(\boldsymbol{\theta})$ determined by trained variational parameters
    $\boldsymbol{\theta}$ and the entanglement gates~\cite{schuld2021a}.

\begin{figure}[t]
		\centering
        \includegraphics{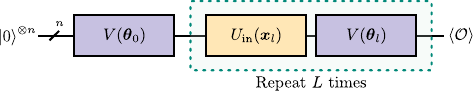}
        \caption{Data re-upload in parameterized quantum circuits:
            After an initial parameterized
			unitary $V(\boldsymbol{\theta}_0)$, $L$ layers of data encoding unitaries
			$U_\text{in}(\boldsymbol{x}_l)$ and parameterized unitaries
			$V(\boldsymbol{\theta}_l)$ are applied to the $n$ qubit quantum register.}%
		\label{fig:reupload}
	\end{figure}

    PQCs with data re-upload have $L$ layers, which consist of data encoding unitaries
    $U_\text{in}(\boldsymbol{x}_l)$ followed by parameterized unitaries
    $V(\boldsymbol{\theta}_l)$ in each layer $l$. As introduced in~\cite{schuld2021a}, the
    circuit starts with parameterized unitaries $V(\boldsymbol{\theta}_0)$ applied on the
    $n$-qubit register $\ket{0}^{\otimes n}$ followed by a sequential implementation of the layers.
    Fig.~\ref{fig:reupload} depicts the circuit layout for such an ansatz.

    We consider two different re-upload strategies resulting in two different implementations of
    $U_\text{in}(\boldsymbol{x}_l)$. In the first case, we follow~\cite{skolik2021} by rescaling
    each continuous classical feature $s_i$ of the state $\boldsymbol{s}$ with trainable parameters
    $\xi_{li}$ for each layer $l$ using the function $x_{li} = \arctan(\xi_{li} s_i) \in
    [-\pi,\pi]$. The index sets of the trainable input parameters are given by $i \in \{1,\ldots,
    n_s\}$ and $l \in \{1,\ldots, L\}$ resulting in $Ln_s$ parameters. In the following,
    $\boldsymbol{x}$ denotes the set of all encoded and rescaled input data and $\boldsymbol{x}_l$ a
    subset of all encoded, and rescaled input data for layer $l$. In this encoding style, each state
    feature $s_i$ is encoded on one qubit:
	\begin{equation}\label{eq:uin1}
	    \boldsymbol{s} \mapsto  U_\text{in,1}(\boldsymbol{x}_l(\boldsymbol{s}))
	        = \bigotimes_{q=1}^n U^{(q)}_{\text{in},1}(x_{lq}),
	\end{equation}
    where $U^{(q)}_{\text{in},1}$ is one of the Pauli rotations $R_x, R_y, R_z$ with rotation angle
    $x_{lq}$ depending on state feature $s_q$ acting on qubit $q$. For this type of encoding the
    number of qubits $n$ has to be equal to the number of input features $n_s$.

    In the second case, we encode, in line with~\cite{perez-salinas2020}, three features of the
    rescaled state $\boldsymbol{s}$ in a universal, single qubit gate $U_{\text{in}, 3}^{(q)}$
    composed of three parameterized Pauli rotation gates. Any combination of rotation gates capable
    of representing a general single qubit rotation, e.\,g. $R_xR_yR_x$, suffices for $U_{\text{in},
    3}^{(q)}$. The state features $s_i$ are encoded as the rotation angles. Therefore, the
    rescaling is done by using different trainable variables $\xi^q_{li}$ for each qubit $q \in
    \{1,\ldots, n\}$, which formally can be written as: $x^q_{li} = \arctan(\xi^q_{li} s_i) \in
    [-\pi,\pi]$. This encoding uses $n L n_s$ trainable input parameters. Similar to the first
    encoding, all trainable parameters used for layer $l$ are denoted by $\boldsymbol{x}_l$. A
    state $\boldsymbol{s}$ is encoded as:
	\begin{equation}\label{eq:uin3}
	    \boldsymbol{s} \mapsto U_\text{in,3}(\boldsymbol{x}_l (\boldsymbol{s})) =
	        \bigotimes_{q=1}^n U_{\text{in}, 3}^{(q)}(x^q_{l1}, x^q_{l2}, x^q_{l3}).
	\end{equation}
    For a state space with more than three features, $U_\text{in,3}$ is repeated until all
    features are encoded. Each $U_\text{in,3}$ then encodes a subset of three features, potentially
    padding the	state space with features set to zero to make its dimensionality divisible  by
    three~\cite{perez-salinas2020}. We perform experiments with both encoding styles, $U_{\text{in},1}
    (\boldsymbol{x})$ and $U_{\text{in},3} (\boldsymbol{x})$, and unify the notation by referring to
    both with $U_\text{in} (\boldsymbol{x})$.

    The parameterized part of the PQC, $V(\boldsymbol{\theta})$, consists of two parts. One
    part includes universal, single qubit gates:
	\begin{equation}\label{eq:upar}
		U_\text{par}(\boldsymbol{\theta}_l) =
        \bigotimes^n_{q=1}U_\text{par}^{(q)}(\theta^q_{l1}, \theta^q_{l2}, \theta^q_{l3}),
	\end{equation}
    which can be implemented by any general, parameterized rotation with three Pauli-rotation gates
    contributing $3 n L$ trainable circuit parameters.
    The second part contains fixed entangling gates $U_\text{ent}$ to create entanglement by acting
    on all $n$ qubits. We choose controlled $Z$ gates on all neighboring pairs of qubits and
    between the last and the first qubit.

    Combining all segments, the PQC ansatz with data re-upload is constructed by applying the
    parameterized part $V(\boldsymbol{\theta}_0)$ to the initial register, followed by a
    layer of data encoding $U_{\text{in}}(\boldsymbol{x}_l)$ and another parameterized part
    $V(\boldsymbol{\theta}_l)$, which are repeated $L$ times. The entire circuit is given by:
    \begin{equation}
        U(\boldsymbol{\theta}, \boldsymbol{x}) =
            \prod_{l=1}^L \Big(
            U_\text{ent}U_\text{par}(\boldsymbol{\theta}_l)U_{\text{in}}(\boldsymbol{x}_l)\Big)
            U_{\text{ent}} U_\text{par}(\boldsymbol{\theta}_0).
    \end{equation}
    This operator is applied to the initial state $\ket{0}^{\otimes n}$ leading to the final state
    $\ket{U(\boldsymbol{\theta},\boldsymbol{x})}$, and the expectation value of an observable
    $\mathcal{O}$:
    \begin{align}
        f_{\mathcal{O}}(\boldsymbol{\theta}, \boldsymbol{s})
            \defeq \langle\mathcal{O}\rangle_{\boldsymbol{\theta}, \boldsymbol{s}}
             = \bra{U\big(\boldsymbol{x}(\boldsymbol{s}), \boldsymbol{\theta}\big)}
         \ \mathcal{O} \
         \ket{U\big(\boldsymbol{\theta}, \boldsymbol{x}(\boldsymbol{s})\big)}.
    \end{align}

    As observables, we choose Pauli-Z gates $\sigma_z^{(1)} \otimes \ldots \otimes \sigma_z^{(n)}$,
    each acting on another qubit to obtain $n$ different output values. The output values can
    either be directly interpreted as values for $Q(\boldsymbol{s}, a)$ in the reinforcement
    learning scenario or combined, scaled, or further post-processed by any classical means including
    additional classical neural network layers.

    Let $a_j$ be one action of the action space $A = \{a_0, \ldots a_{n_a}\}$ with $n_a \leq n_s$.
    If $n_a < n_s$, the PQC output values can either be combined, e.\,g., by multiplying some of
    them~\cite{skolik2021}, to reduce the number of output values to the number of possible actions
    $n_a$, or the first $n_{a}$ qubits are measured. Four our comparative experiments with the
    \texttt{Cartpole-v0} environment we use the former strategy, for the dynamic robot navigation
    environment the latter.

    In our learning scenario, the Q-values range exceeds the interval $[-1, 1]$ and thus needs
    additional post-processing. The authors of~\cite{skolik2021} suggest rescaling each output value
    with an additional trainable parameter $\omega_j$:
    \begin{equation}
        \label{eq:vqc_qfunc}
        Q(\boldsymbol{s},a_j) = \langle\sigma_z^{(j)}\rangle_{\boldsymbol{\theta}, \boldsymbol{s}}
        \cdot w_{j},
    \end{equation}
    which adds $n_a$ trainable output variables to the model.

\section{Method}%
\label{sec:method}
	\subsection{Environments}
        \begin{figure*}[t]
            \centering
            \subfloat[3$\times$3 Environment]{\includegraphics{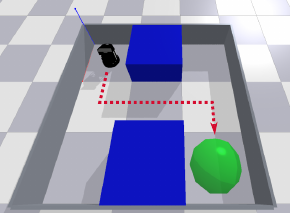}}\qquad
            \subfloat[4$\times$4 Environment]{\includegraphics{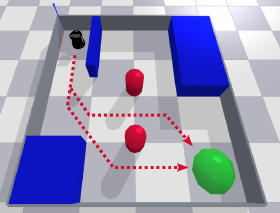}}\qquad
            \subfloat[5$\times$5 Environment]{\includegraphics{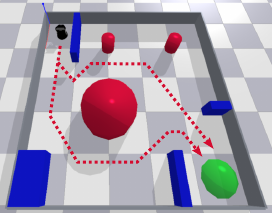}}
            \caption{The three simulated static navigation environments for the
                Turtlebot 2 robot.
                In each, the robot has to navigate from its starting position in the upper left
                corner to the position marked with a green circle in the lower right while avoiding
                collisions with the enclosing walls and any obstacles. With the configured control
                scheme, this takes about 20 steps in the 3$\times$3 environment~(left), 30 in
                the 4$\times$4~(center), and 45 steps in the 5$\times$5 environment~(right)
                for a (near) optimal trajectory. Possible paths the robot
                can take to solve each environment are marked with a red dotted line.}
            \label{fig:envs}
        \end{figure*}

        \begin{figure}[t]
            \centering
            \includegraphics{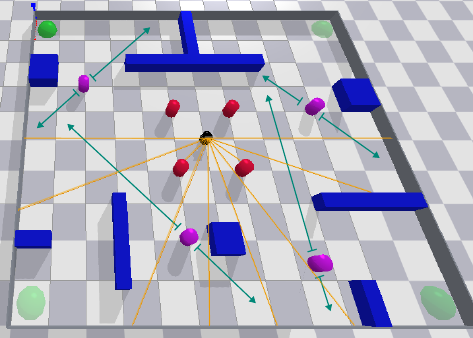}
            \caption{Dynamic environment in which the robot is equipped with a front facing lidar,
                depicted with orange rays. The robot starts in the center of the environment, the
                goal position is sampled at random from either of the four corners at the start.
                While navigating to the goal, the robot has to avoid several static and moving
                obstacles. The trajectories of the moving obstacles is indicated by green arrows.
                Solving the environment takes between 60 and 70 individual steps, depending on the
                sampled goal, position of dynamic obstacles and path the robot takes.}
            \label{fig:env_dyn}
        \end{figure}

        In our experiments, we use four environments based on a simulated Turtlebot\,2
        robot\footnote{https://www.turtlebot.com/turtlebot2/}. We chose this robotic system as it
        enables relatively simple yet realistic navigation tasks while being a readily available
        and extensible system  we can build upon in future work. The robot is controlled via two
        independent motors by setting target velocities for its two wheels.

        The first three environments are static navigation tasks depicted in Fig~\ref{fig:envs}.
        The 3$\times$3 environment shown on the left is the smallest, the 4$\times$4 environment
        (center) is of medium size, and the 5$\times$5 environment (right) is the largest. In each
        environment, the robot starts at a fixed position in the upper left corner and has to
        navigate to a fixed goal position marked with a green sphere while avoiding collisions with
        the outer walls and the obstacles within the environment. The robot has a state space with
        three components for these tasks. The first two are its position in the plain in $s_{x}$ and
        $s_{y}$ coordinates, and the third is its orientation $s_{\varphi}$ around the $z$-axis in
        radians. We use these environments to understand the scaling behavior of parameterized
        quantum circuits in the learning task, assess their behavior and performance for
        trajectories of increasing length and complexity and evaluate the effect of an increasing
        exploration demand.

        We furthermore created a considerably more demanding environment to validate the
        robustness of the presented method with a higher dimensional state space and dynamic
        components in the learning task. In this environment, shown in Fig.~\ref{fig:env_dyn}, the
        robot is equipped with a simple, front-facing lidar that covers a range of
        180\textsuperscript{$\circ$} in the plane. The robot's state space contains ten distance
        measures in 20\textsuperscript{$\circ$} intervals as well as the current distance and
        orientation to the goal.  The robot starts in the center of the environment and has to
        navigate to a goal position, which is sampled at random at the beginning of each episode to
        be in either of the four corners.

        We created all environments with the pybullet~\cite{coumans2016} real-time physics engine
        and set a control frequency of 100\,Hz for collision detection and calculating forward
        dynamics.

        The robot has three actions available (forward, turn left, turn right) to move in the
        environment. To move forward, the same target velocity is applied to both wheels, whereas
        for turning left and right, equal velocities but with opposing directions are set. Turning
        left or right causes a change in orientation between 40\textsuperscript{$\circ$} and
        50\textsuperscript{$\circ$} depending on the current forward and angular velocity of the
        robot. Similarly, the robot moves between $0.15$ and $0.2$ units in the direction of its
        current orientation, where one unit corresponds to the length of one square on the
        environment floor. An action is chosen every 50 simulation steps, corresponding to an
        execution time of 0.5 seconds. With this control scheme, the robot needs about 20
        consecutive actions to reach the goal in the 3$\times$3~environment on a near-optimal
        trajectory, 30 steps in the 4$\times$4~environment, and 45 steps in the large
        5$\times$5~environment. Possible paths the robot can take to solve the static environments
        are marked with red dotted lines in Fig.~\ref{fig:envs}.
        In the dynamic environment, where the robot is equipped with a lidar, a typical trajectory
        leading to the goal takes about 60 to 70 steps, depending on the current goal, position of
        dynamic objects and path taken by the robot.

        We use the same simple yet informative reward function to train the robot in all
        environments. The agent receives a positive reward for decreasing the distance to the goal
        as well as for reaching it, whereas increasing or maintaining the distance as well as
        collisions are penalized. The reward function is given by:
        \begin{equation}
            r(s_{t}, s_{t+1}) = \begin{cases}
                10.0 & \text{if } s_{t+1} \text{ is within the goal area} \\
                \phantom{-}0.1  & \text{if } d_{\text{goal}}(s_{t+1}) < d_{goal}(s_{t}) \\
                -1.0 & \text{for any collision} \\
                -0.2 & \text{else}
            \end{cases}
        \end{equation}
        where $d_{\text{goal}}:S\to\mathbb{R}$ is the euclidean distance of the robot to the goal
        area. The penalties in the reward function ensure that shorter trajectories are preferred by
        the agent. We consider the static environments solved when a total reward of 10.5
        (3$\times$3), 11.0 (4$\times$4), and 10.0 (5$\times$5) is reached. Higher rewards are
        possible, as the two larger environments have more than one possible path to the goal and we
        furthermore allow some tolerance for the length of the trajectory and the exact route.
        Therefore, these thresholds are a lower bound based on several manually determined valid
        trajectories. For the dynamic environment we do not set a threshold and observe the average
        evaluation reward over the entire training time.

        In all environments, an episode ends when the robot reaches the goal, collides with an
        object, or when a maximum of 200 steps were executed during training.

	\subsection{Learning}
        \begin{figure}[t]
            \centering
            \includegraphics{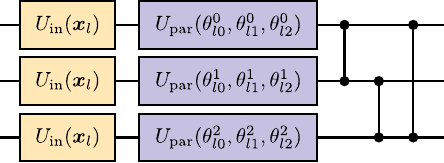}
            \caption{Circuit layout for layer $l$ of our PQC ansatz used for the static environments.
                The encoding unitary $U_{\text{in}}$ on each qubit $q$ is given by
                $U^{(q)}_{\text{in}, 1}$, resp.  $U^{(q)}_{\text{in}, 3}$. The encoding is followed
                by a general rotation gate $U_\text{par}$ with three variational parameters
                $\theta^{q}_{l0}, \theta^{q}_{l1}, \theta^{q}_{l2}$ for each qubit and a full
                entanglement among all qubits with three controlled-$Z$ gates. We omit the qubit
                index on all gates for readability. For the dynamic environment we use the same
                ansatz extended to 12 qubits.}
            \label{fig:qvc_layer}
        \end{figure}

        \begin{figure}[t]
            \centering
            \includegraphics{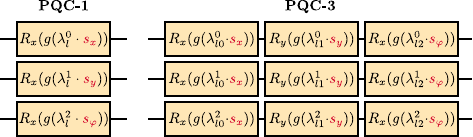}
            \caption{The two input encoding strategies used for training a parameterized
                    circuit as action-value function. For the PQC-1 strategy shown on the left,
                    each feature of the state $\boldsymbol{s}=(s_{x},s_{y},s_{\varphi})$ is encoded
                    on an individual qubit in each layer using a single $R_{x}$ gate. The PQC-3
                    encoding scheme uses three consecutive parameterized rotation gates
                    $R_{x}R_{y}R_{x}$ to encode the entire state $\boldsymbol{s}$ on each qubit and
                    layer as depicted on the right. For both strategies, before encoding, each
                    feature of the state is scaled by a trainable parameter $\lambda_{li}^{q}$
                    individual to each encoding gate, and an activation function
                    $g:\mathbb{R}\to\mathbb{R}$ is applied. Through all our experiments, we use
                    \textit{arctangent} as activation functions.}
            \label{fig:input_enc}
        \end{figure}

        We trained the simulated robot using three different paradigms: A baseline with a classical
        neural network as approximator for the action-value function and two different parameterized
        quantum circuits, distinguished by their encoding strategy for the classical
        input data.

        For the classical baseline agent, we employ a three-layer, fully connected neural network
        with rectified linear unit activation on all but the final layer, which has a linear
        activation. In the static environments, the network takes the three components of the
        robot's state $\boldsymbol{s}=(s_{x}, s_{y}, s_{\varphi})$ as input, followed by two layers
        with $u_{1}$ and $u_{2}$ number of hidden units and three outputs corresponding to
        $Q(\boldsymbol{s}, a_{i}), i\in\{1,2,3\}$. For the dynamic environment, we use the same
        neural network architecture, albeit with a 12 dimensional input for the ten lidar distance
        measurements as well as the distance and orientation to the goal.  The number of trainable
        parameters $|\boldsymbol{\theta}_{NN}|$ including weights and biases for the classical
        neural network is therefore given by:
        \begin{equation}
            |\boldsymbol{\theta}_{NN}| = \underbrace{|s|u_{1} + u_{1}}_{\text{first layer}}
                                  + \overbrace{u_{1}u_{2} + u_{2}}^{\text{second layer}}
                                  + \underbrace{3u_{2} + 3}_{\text{third layer}}.
        \end{equation}
        Here $|s|$ is the dimensionality of the state space and $u_{i}$ the units in the $i$-th
        layer.

        In both quantum cases, we build our circuit on three qubits for the static environments,
        which aligns well with the dimensionality of the state space and the number of actions
        available to the agent. Both circuits follow the general approach depicted in
        Fig.~\ref{fig:reupload} and are only different in their data encoding structure
        $U_\text{in}$ and the number of layers $L$. The circuit layout for a single layer $l>0$ is
        illustrated in Fig.~\ref{fig:qvc_layer}, whereas the encoding strategies are shown in
        Fig.~\ref{fig:input_enc}.

        Our first data re-upload PQC model uses the encoding $U^{(q)}_{\text{in}, 1}$ on each qubit
        with the rotation gate $R_x$ to encode one state feature on each qubit (PQC-1). For the
        second model, we use $U^{(q)}_{\text{in}, 3}$ for each qubit with rotation gates $R_xR_yR_x$
        to encode all three state features on each qubit (PQC-3). As introduced in
        Sec.~\ref{sec:qdrl}, we scale each feature with a trainable parameter that is individual
        for each encoding gate and furthermore apply an activation function for which we choose the
        arctangent in all our experiments. The universal rotation $U^{(q)}_\text{par}$ on each qubit
        is composed of three parameterized Pauli rotation gates $R_xR_yR_z$ with trainable
        parameters.

        For the large, dynamic environment with a 12 dimensional state space, we use circuits with
        12 qubits. The PQC-1 as described above directly translate to this setting, whereas for the
        PQC-3 encoding we distribute all 12 features of the state space across four layers, each
        encoding three of the features, as outlined in Sec.~\ref{sec:qdrl}.

        The number of trainable parameters for each quantum circuit $|\boldsymbol{\theta}_{PQC}|$ is
        the sum of variational parameters in the initial parameterized and the following $L$ layers,
        the input scaling and the output scaling parameters, in total:
        \begin{equation}
            |\boldsymbol{\theta}_{PQC}| = \underbrace{3Q(L+1)}_{\text{variational}}
                                        + \overbrace{Qn_{\text{enc}}L}^{\text{input}}
                                        + \underbrace{3}_{\text{output}}.
        \end{equation}
        Here $n_{\text{enc}}=1$ for the PQC-1 and $n_{\text{enc}}=3$ for the PQC-3 encoding, $Q$ is
        the number of qubits in the circuit.

        Based on these three architectures, two quantum and one classical, we performed experiments
        with different sizes of each architecture for the static environment.  For the classical
        neural network, we evaluated a total of ten configurations for the number of units $u_{1}$
        and $u_{2}$ in the first and second hidden layers. The configurations and their number of
        trainable parameters $|\boldsymbol{\theta}_{NN}|$ are outlined in Table~\ref{tab:conf_nn}.

        \begin{table}[t]
            \centering
            \caption{The configurations used for the classical baseline in all three static
                environments with the number of units
            $u_{1}$ and $u_{2}$ for the first two layers of the neural network and the number of trainable
            parameters $|\boldsymbol{\theta}_{NN}|$ for each configuration.}
            \label{tab:conf_nn}
            \begin{tabular}{ccc}
                \toprule
                \multicolumn{2}{c}{\textbf{Hidden units}} &
                \textbf{Parameters} \\
		        \cmidrule(lr{.5em}){1-2}\cmidrule(lr{.5em}){3-3}
                $\boldsymbol{u_{1}}$ & $\boldsymbol{u_{2}}$ &
                static $\boldsymbol{|\theta_{NN}|}$ \\
		        \cmidrule(lr{.5em}){1-1}\cmidrule(lr{.5em}){2-2}
                \cmidrule(lr{.5em}){3-3}
                8    &  8    &  131     \\
                16   &  8    &  227     \\
                16   &  16   &  387     \\
                32   &  16   &  707     \\
                32   &  32   &  1,238   \\
                64   &  32   &  2,435   \\
                64   &  64   &  4,611   \\
                128  &  64   &  8,963   \\
                128  &  128  &  17,411  \\
                258  &  128  &  34,307  \\
                \bottomrule
            \end{tabular}
        \end{table}
        Likewise, we included ten configurations for each quantum encoding strategy with an
        increasing number of layers $L$. As the different types of encoding lead to a different
        amount of trainable parameters for each layer, we arranged the number of layers to have an
        equal number of parameters between both. The number of layers $L$ for both strategies, as
        well as the number of trainable parameters, including variational, input, and output scaling
        parameters are summarized in Table~\ref{tab:conf_pqc}.

        \begin{table}[t]
            \centering
            \caption{The configurations used for both quantum encoding strategies while training the
            three static environments. The number of layers $L$ as well as the number of trainable
            parameters $|\boldsymbol{\theta_{PQC}}|$, which include the variational, input, and
            output scaling parameters, are outlined.}
            \label{tab:conf_pqc}
            \begin{tabular}{ccc}
                \toprule
                \textbf{PQC-1} & \textbf{PQC-3} & \textbf{Parameters} \\
		        \cmidrule(lr{.5em}){1-1}\cmidrule(lr{.5em}){2-2}
                \cmidrule(lr{.5em}){3-3}
                $L$ & $L$ & static $\boldsymbol{|\theta_{PQC}|}$ \\
		        \cmidrule(lr{.5em}){1-1}\cmidrule(lr{.5em}){2-2}
                \cmidrule(lr{.5em}){3-3}
                12 & 8 & 156 \\
                15 & 10 & 192 \\
                18 & 12 & 228 \\
                21 & 14 & 264 \\
                24 & 16 & 300 \\
                27 & 18 & 336 \\
                30 & 20 & 372 \\
                33 & 22 & 408 \\
                36 & 24 & 444 \\
                39 & 26 & 480 \\
                \bottomrule
            \end{tabular}
        \end{table}
        With regard to the parameter scaling, we emphasize that the number of trainable parameters
        roughly doubles with each increase of the configuration size for the classical baseline,
        whereas the scaling for the quantum circuits is only linear. Thus, the largest neural network
        we employed has about two orders of magnitude (34,307) more parameters than the largest
        quantum circuits (480). With the dynamic environment, we only perform experiments with a
        single configuration for each architecture, due to the considerable computational effort
        involved in simulating very large quantum circuits. The neural network used as baseline has
        256 and 128 hidden units (36,611 trainable parameters), the PQC-1 circuit has 24 layers, and
        the PQC-3 circuit 16 layers (bot 1,191 trainable parameters).

        We set a learning rate for the stochastic gradient descent of $10^{-3}$ for the classical
        baseline as well as for the variational parameters in both quantum
        circuit architectures. The input and output scaling parameters were trained with a learning
        rate of $10^{-2}$ for both PQC-1 and PQC-3 as encoding.

        For the hyper-parameters specific to the DDQN algorithm, we use the same values for all
        experiments. The replay buffer was set to a capacity of 20,000 experience samples and is
        initially filled with 5,000 samples from executing a fully random policy in the environment,
        before each training starts. The agent is trained after each step it executes in the
        environment with a mini-batch of 64 samples from the replay buffer. Exploration is handled
        with an $\epsilon$-greedy policy as introduced in Sec.~\ref{sec:qdrl}, starting at
        $\epsilon=1.0$ and setting $\epsilon\leftarrow0.99\epsilon$ every 250 training steps. Total
        training time is limited to 50,000 steps in all environments, except for the dynamic
        environment, in which we train 100,000 steps.  We evaluate the current performance of the
        learned policy after every 100 training steps by performing 10 consecutive runs within the
        environment. Once the average total reward over those 10 runs surpasses the solution
        criterion for any of the static environment outlined above, the training is stopped early,
        whereas we do not stop the training early in the dynamic environment.

        To gather sufficient data on the robustness and reproducibility of the learning procedure,
        we repeat the training for each combination of static environment, architecture, and
        configuration 20 times, each time with a different random seed. We do not set the random
        seeds to specific values but have them provided by the operating system's randomness source
        instead.  We consider a configuration successful if at least 15 of 20 training runs solve
        the environment. In the dynamic environment, we repeat each training 10 times and record the
        evaluation performance to evaluate the robustness of the presented methods in a considerably
        larger and more challenging environment and large quantum circuits.

        All hyper-parameters were determined empirically before the actual experiments. Our main
        goal was to find a set of parameters that would enable reliable and robust training
        under mostly identical premises for all three architectures, their respective configurations,
        and for all three environments, as our main interest is not in absolute performance but in
        comparison of architectures and scaling behavior. An overview of all
        hyper-parameters can be found in Table~\ref{tab:hyper} in App.~\ref{sec:app_hyperp}.

        \subsection{Hardware, Software and Computational Resources}
        All experiments were conducted on a workstation equipped with an AMD Ryzen
        Threadripper Pro 3975WX 32 core/64 thread CPU, 128 GB of RAM, and an NVIDIA RTX A6000 GPU\@.
        On the software side, we used TensorFlow~\cite{tensorflow2015} as framework for all general
        and classical machine learning tasks, TensorFlow Quantum~\cite{broughton2021} for quantum
        machine learning specific tasks, as well as TensorFlow Agents~\cite{TFAgents2018} for all
        components related to Deep Reinforcement Learning and a stable DDQN implementation.
        TensorFlow Quantum integrates the Cirq~\cite{cirq2022} quantum computing framework for
        building and running quantum circuits, as well as the Qsim~\cite{qsim2020} quantum circuit
        simulator.

        All quantum circuit simulations in Qsim were executed under idealized noise-free conditions.
        We compute the expected value of observables directly from the system's state vector.  If
        experiments were to be conducted in a shot-based simulation, a large enough number of
        circuit receptions would need to be chosen to estimate the expected values of observables
        with sufficient accuracy. Similarly, if experiments were to be reproduced on quantum
        hardware or with simulated hardware noise, appropriate measures for error mitigation would
        have to be taken into account, which is outside of the scope of this work.

        All simulated robotic environments were built using the PyBullet~\cite{coumans2016} python
        bindings to the Bullet real-time physics SDK\@. For the baseline experiments described in
        App.~\ref{sec:app_baseline}, we furthermore used the OpenAI Gym~\cite{brockman2016} suite.

        We released our robotic environments as well as the entire experimental setup under an Open
        Source license for interested researchers to reproduce, verify, or build upon our work. Both
        can be found together with installation and usage instructions under the following addresses:

        \begin{itemize}
            \item Environments: https://github.com/dfki-ric-quantum/qdrl-turtlebot-env
            \item Experiments: https://github.com/dfki-ric-quantum/qdrl-turtlebot-eval
        \end{itemize}

        Concerning the computational resources and wall-clock time needed to conduct our
        experiments, we observe the following: For the classical baseline, training a single neural
        network within the range of configurations and across all environments requires 1.5 GB of
        RAM and 600 MB of VRAM, assuming TensorFlow uses GPU acceleration. Training the network for
        1.000 steps takes on average 25 seconds wall-clock time with our hardware setup, which is
        relatively stable overall environments and network sizes.

        For the three static environments, the number of qubits of a quantum circuit, which is the
        same in both our encodings and across all configurations, primarily determines the memory
        requirements for its simulation. Simulating the training of each quantum circuit requires
        about 2.2 GB of system memory and 500 MB of VRAM. The quantum circuit simulator imposes a
        substantial computational overhead, resulting in much longer execution in terms of
        wall-clock time and a nearly linear growth with respect to the number of layers. The average
        wall-clock time for 1.000 training steps in the 5$\times$5 environment with the PQC-1 and
        PQC-3 encoding are summarized in Table~\ref{tab:wc_pqc}.
        \begin{table}[ht]
            \centering
            \caption{Average wall-clock time for 1.000 training steps with the PQC-1 and PQC-3encoding
            in the 5$\times$5 environment. The time necessary to train the model grows nearly linear
            in the number of layers.}
            \label{tab:wc_pqc}
            \begin{tabular}{ccccc}
                \toprule
                \multicolumn{2}{c}{\textbf{PQC-1}} & &
                \multicolumn{2}{c}{\textbf{PQC-3}} \\
                \cmidrule(lr{.5em}){1-2}
                \cmidrule(lr{.5em}){4-5}
                $\boldsymbol{L}$ & \textbf{Wall-clock time (s)} & &
                $\boldsymbol{L}$ & \textbf{Wall-clock time (s)} \\
                \cmidrule(lr{.5em}){1-1}
                \cmidrule(lr{.5em}){2-2}
                \cmidrule(lr{.5em}){4-4}
                \cmidrule(lr{.5em}){5-5}
                12 & 393  & & 8 & 361 \\
                15 & 467  & & 10 & 442 \\
                18 & 544  & & 12 & 520 \\
                21 & 629  & & 14 & 605 \\
                24 & 722  & & 16 & 661 \\
                27 & 815  & & 18 & 743 \\
                30 & 878  & & 20 & 811 \\
                33 & 967  & & 22 & 892 \\
                36 & 1033 & & 24 & 954 \\
                39 & 1088 & & 26 & 994 \\
                \bottomrule
            \end{tabular}
        \end{table}
        Learning the dynamic environment with either encoding on 12 qubits for the number of layers
        we use, requires considerably more computational resources. A single run requires
        about 18 GB of system memory and 1.2 GB of VRAM. Training for 1.000 steps takes on average
        one hour, hence the training the full 100.000 steps is finished in about four days.

\section{Results}%
\label{sec:results}

    \begin{figure*}[t]
        \centering
        \includegraphics{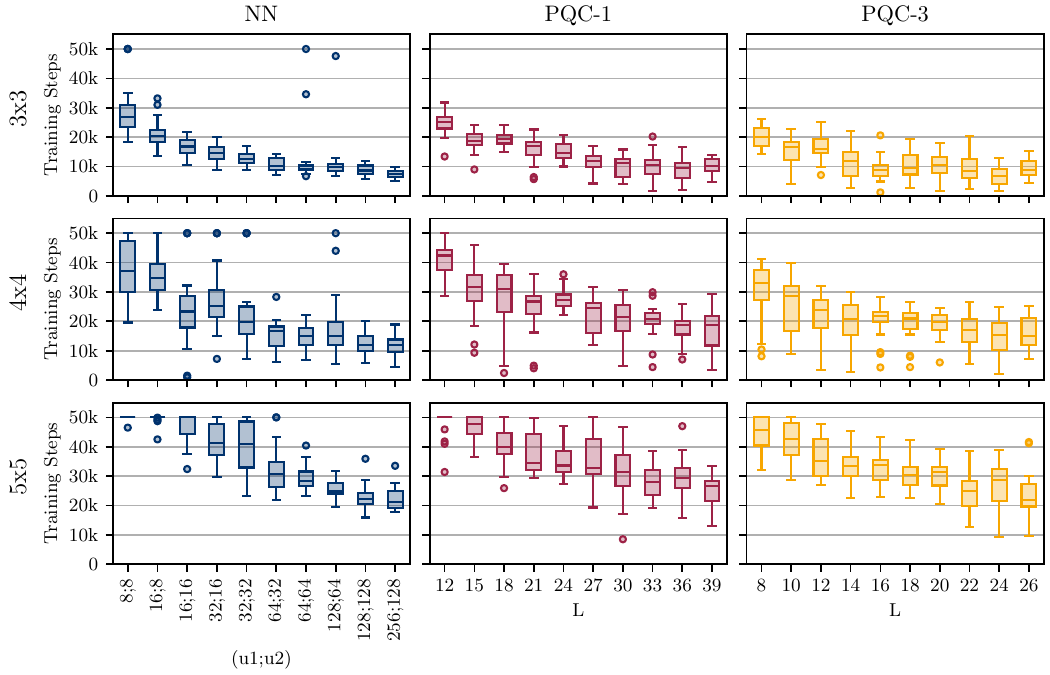}
        \caption{Statistics on training time for all three static environments, architectures, and
        configurations. The results per environment are shown from the top to bottom row, whereas the
        classical baseline neural network architecture (NN), as well as both types of quantum circuits
        (PQC-1 and PQC-3) are arranged from left to right. For each combination of environment and
        architecture all related configurations, that is number of units $(u_{1},u_{2})$ for the classical
        baseline and number of layers $L$ for both quantum encoding strategies, are
        reported. Each box shows the median training steps over 20 runs for each configuration with
        different random seeds as well as the lower and upper quartile, range and flier points.}
        \label{fig:box}
    \end{figure*}

    \begin{figure*}[t]
        \centering
        \includegraphics{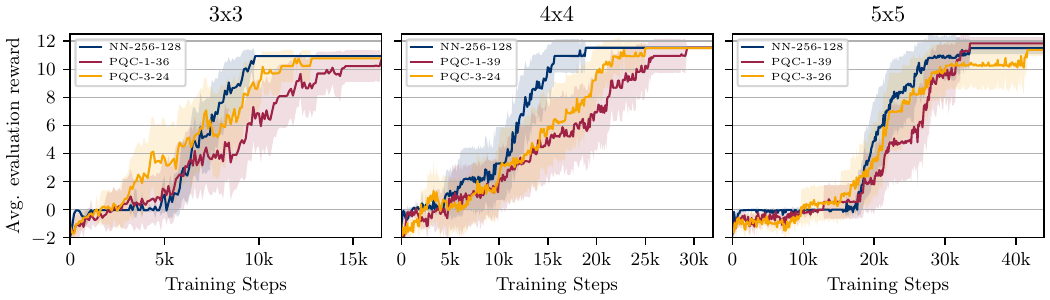}
        \caption{Evaluation results for the configurations with the best mean performance
            for each architecture in all three static environments.  From left to right the mean
            evaluation performance over 20 consecutive runs with the trained policy outlined by the
            95\% confidence interval for the $3\times 3$, $4\times 4$, and $5\times 5$ environments
            is shown.  Negative rewards are rescaled by a factor of $0.1$ to improve readability.
            When not bound by the number of trainable parameters, the classical baseline models
            performs slightly better than both quantum architectures in all three environments.
            Training was stopped once the mean evaluation reward reached the solution threshold in
            the environment, plots are padded with additional evaluation runs for better comparability.}
        \label{fig:best_20}
    \end{figure*}

    \begin{figure*}[t]
        \centering
        \includegraphics{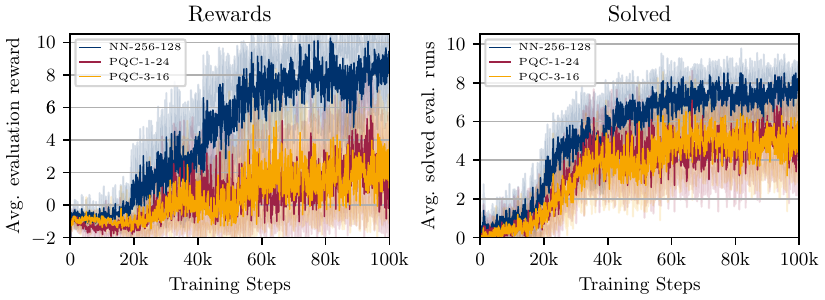}
        \caption{Evaluation performance for the three tested configurations in the dynamic
        environment with regards to the evaluation rewards (left) and number of solved evaluation
        runs (right). Both metrics are recorded every 100 training steps for 10 consecutive
        evaluation runs and 10 receptions of the experiment with different random seeds. The
        outlines mark the 95\% confidence interval. Negative rewards are scaled by a factor of $0.1$
        to improve readability. For both metrics we find, that while there is learning progress for
        both quantum circuit architectures, the classical baseline provides better and more robust
        results, albeit with about 30 times the number of trainable parameters.}
        \label{fig:dyn_res}
    \end{figure*}

	\subsection{OpenAI Gym environments}%
	\label{sec:results_openaigym}
    As we work with custom environments, we first compared their complexity to established
    OpenAI Gym environments, namely \texttt{FrozenLake} and \texttt{Cartpole-v1}. The results for
    both environments with classical neural networks and PQCs are described in
    Appendix~\ref{sec:app_baseline}. With these comparative experiments, we can demonstrate that our
    navigation environments are indeed substantially more difficult to solve for the DDQN algorithm.

	\subsection{Static environments}%
	\label{sec:results_static}
    Performing experiments with the 10 classical neural network configurations and 10 configurations
    for both PQC input encoding variants provides insight into the scaling behavior and robustness
    across multiple training runs for each architecture in the given robotic reinforcement learning
    task. The complete statistics for all experiments in the static environments are outlined in
    Tables~\ref{tab:res3} to~\ref{tab:res5} in Appendix~\ref{sec:app_res} and visualized in
    Fig.~\ref{fig:box}.

    The first noteworthy result is that all three architectures, the classical neural network as
    well as both types of quantum circuits are capable of learning an optimal action-value function
    in all three environments in 20 out of 20 training runs with a sufficiently large
    configuration (see column \textit{Solved} in the Tables~\ref{tab:res3} to~\ref{tab:res5}). More
    precisely, for the $3\times 3$ and $4\times 4$ environments, all configurations of the
    architectures solve the environments, whereas in the $5\times 5$, the three smallest neural
    networks, the two smallest PQC-1, and the smallest PQC-3 configurations were unable to solve the
    environment in at least $15$ out of $20$ runs. Also, we find that an increase of the model size
    in terms of the number of trainable parameters leads to a decreased median and mean in required
    training steps. This trend converges after the model size reaches a sufficient size. In our
    case, the two biggest configurations of all three architectures are similar in terms of median
    and mean of training steps. In addition, in most cases, the range and standard deviation
    decreases as the model size increases, resulting in our largest models being the best-performing
    and most stable configurations.

    In the following, we focus on the two best PQC-1 and PQC-3 configurations and compare them with
    four different neural network configurations. From our data, we select the classical neural
    networks such that they have a similar number of parameters or one or two orders of
    magnitude more parameters than the PQC configurations. Table~\ref{tab:scaling} summarizes the
    mean and standard deviation of the required training steps for these configurations. From this,
    we observe the following general trends with regard to the number of trainable parameters:

    \begin{itemize}
        \item With about the same order of magnitude of parameters, the quantum circuits perform
            better and converge to an optimal solution faster. This is especially true for the PQC-3
            architecture.
        \item With about one order of magnitude more parameters for the classical neural network,
            its performance is about equal compared to the parameterized quantum circuits.
        \item A further increase in the number of parameters up to two orders of magnitude more for
            the neural network puts it slightly ahead of both quantum circuit architectures in all
            observed metrics.
    \end{itemize}

    \begin{table*}[t]
        \caption{Mean number of training steps and standard deviation in all three environments for the
            two largest configurations for both quantum circuit architectures in comparison to two
            classical baseline models with about the same order of magnitude of trainable parameters
            as well as two larger neural networks. We find, that with about the same order of
            magnitude of parameters, the two quantum architectures converge to an optimal solution in
            fewer training steps. With an order of magnitude more parameters, the classical neural
            network performs comparable or better and achieves better performance compared to
            both quantum architectures with further increasing number of trainable parameters.}
        \label{tab:scaling}
        \centering
        \begin{tabular}{ccccccccc}
            \toprule
            & & & \multicolumn{6}{c}{\textbf{No.\ of training steps}} \\
            \cmidrule(lr{.5em}){4-9}
            & \multicolumn{2}{c}{\textbf{Environment:}} &
            \multicolumn{2}{c}{$\boldsymbol{3\times 3}$} &
            \multicolumn{2}{c}{$\boldsymbol{4\times 4}$} &
            \multicolumn{2}{c}{$\boldsymbol{5\times 5}$} \\
            \cmidrule(lr{.5em}){2-3}
            \cmidrule(lr{.5em}){4-5}
            \cmidrule(lr{.5em}){6-7}
            \cmidrule(lr{.5em}){8-9}
            \textbf{Arch.} & \textbf{Config.} & $\boldsymbol{|\theta|}$ & \textbf{Mean} &
            \textbf{Std.} & \textbf{Mean} & \textbf{Std.} & \textbf{Mean} & \textbf{Std.} \\
            \cmidrule(lr{.5em}){1-1}
            \cmidrule(lr{.5em}){2-2} \cmidrule(lr{.5em}){3-3} \cmidrule(lr{.5em}){4-4}
            \cmidrule(lr{.5em}){5-5} \cmidrule(lr{.5em}){6-6} \cmidrule(lr{.5em}){7-7}
            \cmidrule(lr{.5em}){8-8} \cmidrule(lr{.5em}){9-9}
            \multirow{4}{*}{\textbf{NN}}
            & (16;8)    & 227    & 21,075 & 5,050 & 35,910 & 7,672  & 49,515 & 1,684 \\
            & (16;16)   & 387    & 16,405 & 3,129 & 24,565 & 13,736 & 46,570 & 5,237 \\
            & (64;32)   & 2,435  & 10,635 & 2,290 & 15,055 & 4,315  & 32,135 & 8,049 \\
            & (256;128) & 34,307 & 7,495  & 1,359 & 11,480 & 3,899  & 22,220 & 4,122 \\
            \midrule
            \multirow{2}{*}{\textbf{PQC-1}}
            & 36        & 444    &  9,150  & 3,746 & 17,705 & 4,678 & 29,260 & 6,614 \\
            & 39        & 480    & 10,060  & 2,905 & 16,880 & 7,288 & 25,110 & 5,309 \\
            \midrule
            \multirow{2}{*}{\textbf{PQC-3}}
            & 24        & 444    &  6,850  & 3,146 & 14,665 & 6,068 & 25,757 & 7,334 \\
            & 26        & 480    &  9,515  & 3,347 & 15,875 & 5,164 & 23,635 & 7,535 \\
            \bottomrule
        \end{tabular}
    \end{table*}

    Furthermore, we can compare the results of the two best PQC-1 and PQC-3 configurations in
    Table~\ref{tab:scaling}. For all environments, the best PQC-3 architecture yields faster
    convergence to an optimal policy compared to the best PQC-1 encoding scheme.

    This advantage is relatively stable across all three static environments, suggesting that for
    the navigation setup considered in this work, a larger number of encoding gates is beneficial. A
    larger variety of environments with regards to complexity and type of task to learn would need
    to be evaluated to make more definitive statements on this.

    The evaluation performance for the best configuration for each architecture in all three
    environments is shown in Fig.~\ref{fig:best_20}. During training, we observed the agent's
    performance with the trained policy every $100$ steps for $10$ consecutive runs and evaluated
    its mean reward. In all three environments, the classical baseline converges to an optimal
    policy faster, albeit with two orders of magnitude more trainable parameters. In the $3\times 3$
    environment, the PQC-3 architecture reaches a solution notably faster than the PQC-1
    architecture. With increasing environment complexity, both types of quantum circuits perform
    increasingly similarly, whereas the classical neural network remains ahead of both. This finding
    emphasizes the trends discussed above.

	\subsection{Dynamic environment}%
	\label{sec:results_dyn}
    In the large, dynamic navigation environment, our main interest is the robustness of the
    presented method in a substantially more demanding task and employing considerably larger
    quantum circuits. To this end, we trained a classical baseline and two large quantum circuits
    with the two encoding strategies for 100,000 iterations on the environment. We evaluated the
    performance in 10 consecutive runs every 100 training steps. Fig.~\ref{fig:dyn_res} shows the
    training progress regarding the mean evaluation reward and number of solved evaluation runs, with
    averages taken over 10 repetitions of the experiment with different random seeds.

    The classical baseline neural network performs considerably better in this task
    than both employed quantum circuits, learns policies that achieve higher mean rewards, solves
    more evaluation runs on average and is more robust in the dynamic setting. Both quantum
    architectures perform about the same concerning to both metrics. The fact that the difference
    between the classical model and quantum circuits regarding the mean reward is larger than for
    the number of solved evaluation runs is explained by the observation that the environment
    allows for much larger negative rewards on failed runs than positive rewards on the successful
    ones. Hence, the negative rewards will dominate the result if several runs fail.

    Table~\ref{tab:dyn_stats} summarizes the best results achieved by all three architectures. The
    classical baseline reaches its best average performance after 81,500 training steps, whereas
    both quantum circuits require more than 94,000 steps. Additionally, the mean evaluation
    reward of 10.27 for the classical neural network is considerably larger than 5.50 and 3.87 for
    the PQC-1 and PQC-3 architecture.

    \begin{table}[ht]
        \centering
        \caption{Statistics over the training in the dynamic navigation environment. For all three
        configurations the number of training steps to the best performing evaluation runs, the
        mean evaluation reward, the mean number of solved evaluation runs as well as their
        respective standard deviations are reported. Statistics are taken over 10 consecutive
        evaluation runs executed every 100 steps and 10 repetitions of the experiment with different
        random seeds.}
        \label{tab:dyn_stats}
        \begin{tabular}{ccccccc}
            \toprule
            & & &
            \multicolumn{2}{c}{\textbf{Reward}} &
            \multicolumn{2}{c}{\textbf{Solved}} \\
            \cmidrule(lr{.5em}){4-5}
            \cmidrule(lr{.5em}){6-7}
            \textbf{Arch.} & \textbf{Config.} & \textbf{Steps} &
                \textbf{Mean} & \textbf{Std.} & \textbf{Mean} & \textbf{Std.} \\
            \cmidrule(lr{.5em}){1-1}
            \cmidrule(lr{.5em}){2-2}
            \cmidrule(lr{.5em}){3-3}
            \cmidrule(lr{.5em}){4-4}
            \cmidrule(lr{.5em}){5-5}
            \cmidrule(lr{.5em}){6-6}
            \cmidrule(lr{.5em}){7-7}
            \textbf{NN}    & (256;128) & 81,500 & 10.27 & 1.37 & 8.50 & 0.92 \\
            \textbf{PQC-1} & 24        & 94,200 & 5.50  & 3.12 & 6.70 & 1.79 \\
            \textbf{PQC-3} & 16        & 94,500 & 3.87  & 3.53 & 6.30 & 1.27 \\
            \bottomrule
        \end{tabular}
    \end{table}

    After this training duration, the robot can successfully navigate to the goal on average
    in 8.5 out of 10 evaluation runs over 10 repeated experiments. Solving 6.7 and 6.3 evaluation
    runs on average for the quantum architectures shows noteworthy training progress for both, but with
    considerably worse performance. We furthermore observe larger standard variations on both
    metrics for the quantum models compared to the classical baseline, suggesting less robust and
    less reliable training results.

\section{Discussion}%
\label{sec:discussion}
    In this work, we investigated the potential and scaling of hybrid quantum deep
    reinforcement learning as a method to learn autonomous robotic behaviors. We systematically
    evaluated two different quantum circuit architectures in three simulated static environments of
    increasing difficulty and with increasing circuit sizes. These results were compared to a
    classical neural network baseline. Additionally we tested the robustness of the presented method
    in a considerably more demanding learning task, using a dynamic navigation environment.

    Both quantum architectures as well as the classical baseline yielded sufficient action-value
    functions for the simulated robot in all three static environments. Not considering the
    number of trainable parameters, the classical baseline models outperformed the quantum circuits
    in terms of training speed and stability. A noteworthy result, which is in line with previous
    findings from the quantum deep reinforcement learning research is that both best-performing
    quantum circuits were capable of solving the environments within a similar number of training
    steps as classical neural networks with about one order of magnitude more trainable parameters.
    This observation is consistent across all three environments. The best-performing quantum models
    have $444$ and $480$ trainable parameters, the classical baseline was sufficient to solve the
    $3\times 3$ and $4\times 4$ with a similar amount of parameters, albeit with substantially more
    training steps. At this model size, the neural network was unable to fit an optimal
    action-value function for the $5\times 5$ environment in most of the $20$ training runs
    within the $50,000$ training step threshold we set, whereas both quantum architectures still
    succeeded with only $300$ parameters.

    Comparing both quantum circuit architectures, we find that the PQC-3 embedding performs better
    than the PQC-1 embedding in all three environments, suggesting that in this context, having more
    encoding gates for the same data is beneficial, although the difference becomes less pronounced
    with increasing environment difficulty.  Moreover, our experiments show that with increasing
    environment size, quantum circuits with more layers are needed to solve the tasks consistently,
    especially for the $5\times 5$ environment.  This finding is consistent with the results
    from~\cite{schuld2021a}, as adding more layers increases the expressiveness of the circuit,
    which makes it possible to approximate more complex action-value functions.

    Testing the same learning methods in a more demanding, dynamic navigation environment, we find
    that both quantum circuit architectures get outperformed by the classical neural network
    regarding reward, solved evaluations, training duration and robustness. Given the limited scope
    of this experimental setup, it remains open, if this result can be improved by changes on the
    training procedure, circuit architecture, encoding strategies or by increasing the circuit size.
    We consider these questions to be out of scope for this work, but plan to address them in
    future.

    Additionally, our results demonstrate that PQCs of this size are trainable in a quantum
    circuit simulator for a practical problem class, which does not necessarily follow from previous
    considerations on the expressiveness of PQCs~\cite{perez-salinas2020, schuld2021a}.  Beyond
    these results, we can confirm, similar to e.\,g.,~\cite{Franz2022}, that training PQCs is fairly
    unstable regarding changes in the hyperparameters compared to classical neural networks.

    Considering the best-performing PQCs architectures in this work, we have to emphasize that this
    configuration is not to be considered efficient or even viable for current quantum hardware.
    The largest employed circuit using the PQC-3 architecture has almost 200 gates per qubit, not
    considering additional gates that could be introduced by transpiling it to a native gate set of
    any quantum hardware platform. Circuits with long execution times and more gates are more prone
    to noise on current quantum hardware. Hence, we would not expect meaningful results without
    substantial error mitigation efforts. Training the circuits directly on quantum hardware was
    also not a realistic option, given the total number of experiments we conducted and the limited
    availability and access to quantum computing hardware. Consequently, we limited our study to an
    idealized environment in a noise-free quantum circuit simulator.

\section{Outlook}%
\label{sec:outlook}
    \begin{figure*}[t]
        \centering
        \includegraphics{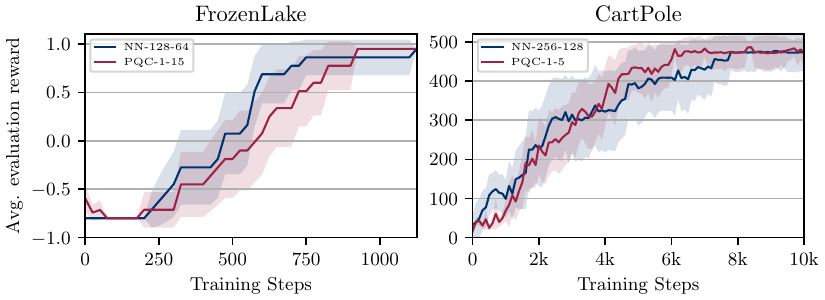}
        \caption{Evaluation results of the FrozenLake and CartPole environment.
            Each plot shows the mean evaluation performance over 20 consecutive runs with the
            trained policy. The evaluation was performed every 25, resp. 100, training steps and
            training stopped, once the environment was solved with the policy. Plots are padded with
            additional evaluation runs to have equal lengths for better comparability. The outline
            of each plot is the 95\% confidence interval.
        }
        \label{fig:benchmark_results}
    \end{figure*}

    Understanding the characteristics of PQCs is an ongoing research topic. For PQCs to offer
    advantages over classical solutions, there are still some open questions that have to be
    addressed. Concerning expressiveness, the authors of~\cite{schuld2021a} showed that PQCs with
    the data-reupload technique can represent real-valued truncated Fourier series.
    While this could be considered a weak restriction on the expressiveness, it remains unclear
    if they are rich enough to approximate deep RL algorithm outputs for more complex behaviors.
    In~\cite{schreiber2022}, the authors leverage that PQCs represent truncated Fourier
    series by showing that classical models can be obtained efficiently from trained PQCs. They
    also report no advantage in the performance nor trainability of PQCs over classical models for the
    problems they consider. The trainability of PQCs is analyzed in more detail by Bittel et
    al.~\cite{bittel2021}, who rigorously prove that classical training is NP-hard, and by the authors
    of~\cite{anschuetz2022}, who found many sub-optimal local minima in the gradient landscape.
    Moreover, barren plateaus~\cite{mcclean2018} are one additional hurdle for trainability. These
    works and our results indicate that PQCs mark the beginning of quantum machine learning in
    general and quantum deep reinforcement learning specifically. These methods have to be developed
    further substantially to yield potential improvements over classical learning techniques.

    Our results provide additional insight into the scaling behavior and applicability of hybrid
    quantum deep reinforcement learning based on PQCs, especially with regard to more demanding
    problems than previously considered.  Our experimental setting is focused on three static
    environments and two different quantum circuit architectures. Furthermore, we studied the
    robustness of these methods in a more demanding, dynamic navigation task, although with limited
    scope. Hence more empirical research is needed to substantiate our findings further, and produce
    more conclusive results.

    The second area is the applicability of quantum machine learning and quantum deep reinforcement
    learning in real-world applications, especially in the field of robotics. While we have
    demonstrated quantum deep reinforcement learning in a limited robotic scenario, actual
    advantages of the presented method over classical deep reinforcement learning have yet to be
    shown. While previous works demonstrated a quantum advantage for a certain class of
    problems~\cite{jerbi2021} intractable for classical learning methods, it remains an open
    question if this advantage can be translated to problems from e.\,g., robotic domains.

    Another crucial topic linked to real-world applications, is the encoding scheme of classical
    data into the quantum circuit. With the proposed methods, the required number of qubits and the
    operations per qubit scale linearly in the best case with the dimensionality of the state
    space. It will be interesting to see how different encoding techniques like e.\,g., amplitude
    encoding~\cite{larose2020} would impact the learning behavior. Also, we limited our
    experiments to state spaces of small dimensionality to account for the computational demands of
    simulating quantum circuits on a classical computer. While this imposed no detriments on our
    learning scenarios, having high dimensional sensory data, e.\,g., high-resolution image data, is
    common in more complex robotic tasks. How to encode classical data with hundreds, thousands, or
    more dimensions efficiently onto quantum devices with their current limitations is an open
    question. Investigating classical pre-processing, compression, and dimensionality reduction
    techniques in this context could potentially enable quantum deep reinforcement learning for such
    scenarios. Tensor networks, as indicated by Chen et al.~\cite{chen2022}, are one further promising
    candidate for encoding more complex robotic data.

    Additionally, the scope of our work is limited with regards to actual quantum hardware and its
    properties. We performed all our experiments with a quantum circuit simulator which, enabled us
    to employ circuits with a depth beyond what current hardware provides and also removed
    the necessity to deal with the noise that typically comes with the execution of algorithms on
    current quantum hardware. The execution of even simpler machine learning tasks on actual quantum
    hardware would be further limited by their sparse availability, complexity, and high usage cost.
    To circumvent these issues, research into techniques to combine quantum simulators and quantum
    hardware in an efficient training setup could be a practical route forward.

    We understand our work as a contribution toward application-focused empirical research on
    quantum algorithms in a robotic context. We see this as a viable route to accelerate the
    development and understanding of quantum algorithms, quantum machine learning, and
    the application of quantum techniques in deep reinforcement learning. Looking forward, we
    believe that quantum algorithms, together with future hardware developments in the field of
    quantum computing, will contribute to the advancement of autonomous robotics.

\section{Acknowledgment}
\label{sec:ack}
    We thank Elie Mounzer, Gunnar Schönhoff, Patrick Draheim, Melvin Laux, and Alexander Fabisch for
    their valuable feedback on this manuscript. We furthermore thank the anonymous reviewers for
    their constructive remarks and recommendations to improve this paper.

\appendices

\section{\break Comparison to baseline environments}
\label{sec:app_baseline}

    We use our learning setup to solve the benchmark OpenAI gym~\cite{brockman2016} environments
    \texttt{FrozenLake} and \texttt{CartPole-v1}. This way, we underline our argument that the
    navigation tasks are indeed more complex and difficult to learn.

    To learn the \texttt{FrozenLake} environment, we use binary encoding for the state features,
    adapt the circuit to four qubits, and adapt the parameters for epsilon decay and max steps per
    episode. All other learning hyper-parameters are unchanged. We also use the \textit{arctangent}
    activation function and trainable parameters on the input features as well as four trainable
    output parameters. A full list of the hyper-parameters is given in Table~\ref{tab:hyper} in
    App.~\ref{sec:app_hyperp}. The left plot of Fig.~\ref{fig:benchmark_results} shows that 20
    runs with a classical neuronal networks with $(128,64)$ hidden units learn an optimal policy in
    fewer than 1,250 training steps with a mean of 510 steps, a median of 513 steps, and a standard
    deviation of 204 steps. The classical architecture takes roughly 20 times as long to learn an
    optimal policy in our simplest navigation task. Similar results hold for PQCs with one input
    encoding and 15 layers (PQC-1-15). Here, the training finishes on average in 593 steps, with a
    median of 613 steps and a standard deviation of 205 steps. This result is in alignment
    (slightly better) with~\cite{skolik2021}. We want to emphasize that we did not fine-tune the
    hyper-parameters for the \texttt{FrozenLake} environment but were still able to learn the task
    much faster than for our $3\times 3$ navigation environment.

    We obtained similar results for the \texttt{Cartpole-v1} environment as depicted in the right
    plot of Fig.~\ref{fig:benchmark_results}. For this environment, we adapted the PQC to 4 qubits
    and used the measurements $\sigma^{(1)}_z\sigma^{(2)}_z$ and $\sigma^{(3)}_z\sigma^{(4)}_z$ for
    the post-processing. We adapted the epsilon decay parameters and other hyper-parameters
    slightly, as shown in Table~\ref{tab:hyper}. Averaged over 20 runs, the classical network with
    $(256, 128)$ hidden units is able to solve \texttt{CartPole-v1} with an average of 3,645 training
    steps (median: 2,350, standard deviation: 2,600) with slightly adapted hyper-parameters. That is
    approximately twice as fast as the same network architecture learns the $3\times 3$ navigation
    environment. For the PQCs, the configuration with one input encoding and five layers needed
    fewer than 10,000 training steps to learn the optimal policy, which is notably faster than
    reported in the literature (e.\,g.,~\cite{skolik2021} for \texttt{Cartpole-v0}). The PQC-1-5
    ansatz solves \texttt{CartPole-v0} in a similar time (mean: 4065, median: 4050, standard
    deviation: 1933) and thus solves it faster than larger PQC-1 configurations solve the $3 \times
    3$ navigation environment.

    Hence, we conclude that our navigation tasks are considerably more challenging than
    \texttt{FrozenLake} and \texttt{Cartpole-v1} for the (hybrid quantum) DDQN algorithm.

\section{\break Hyper-parameters for experiments}
\label{sec:app_hyperp}

    The hyper-parameters used in all environments and learning setups are outlined in
    Table~\ref{tab:hyper}.

\section{\break Result Details}
\label{sec:app_res}

    Detailed statistics over all conducted experiments are reported in
    Tables~\ref{tab:res3},~\ref{tab:res4} and~\ref{tab:res5}.

\bibliographystyle{unsrt}
\bibliography{main}

\clearpage
\onecolumn

\begin{table}[ht]
	\centering
	\caption{
        Summary of all hyper-parameters used in the training of the four simulated robotic
        environments and the two OpenAI gym environments used for reference and comparison in
        App.~\ref{sec:app_baseline}.
	}
	\label{tab:hyper}
	\begin{tabular}{rrrrrrr}
		\toprule
		\textbf{Hyper-Parameter} & \textbf{3$\times$3} & \textbf{4$\times$4} & \textbf{5$\times$5}
		 & \textbf{dynamic} & \textbf{FrozenLake} & \textbf{Cartpole-v1} \\
		\cmidrule(lr{.5em}){1-1}
		\cmidrule(lr{.5em}){2-2}
		\cmidrule(lr{.5em}){3-3}
		\cmidrule(lr{.5em}){4-4}
		\cmidrule(lr{.5em}){5-5}
		\cmidrule(lr{.5em}){6-6}
		\cmidrule(lr{.5em}){7-7}
        Max.\ training steps & 50,000 & 50,000 & 50,000 & 100,000 & 2,000 & 10,000\\
        Max.\ steps per episode & 200 & 200 & 200 & 200 & 80 & 500 \\
        Initial random steps & 5,000 & 5,000 & 5,000 & 5,000 & 5,000 & 5,000 \\
		\midrule
        Eval.\ runs & 10 & 10 & 10 & 10 & 10 & 100\\
        Eval.\ every $t$ steps & 100 & 100 & 100 & 100 & 25 & 100 \\
        Eval.\ threshold & 10.5 & 11 & 10 & -/- & 0.95 & 475 \\
		\midrule
        $\epsilon$ starting value & 1.0 & 1.0 & 1.0 & 1.0 & 1.0 & 1.0 \\
        $\epsilon$ minimum & 0.1 & 0.1 & 0.1 & 0.1 & 0.1 & 0.05 \\
        $\epsilon$ decay & 0.99 & 0.99 & 0.99 & 0.99 & 0.9 & 0.95 \\
        Decay every $t$ steps & 250 & 250 & 250 & 500 & 50 & 100 \\
		\midrule
        Replay buffer size & 20,000 & 20,000 & 20,000 & 20,000 & 20,000 & 20,000 \\
        Batch size & 64 & 64 & 64 & 64 & 64 & 64 \\
        Target update & 100 & 100 & 100 & 100 & 100 & 10 \\
        $\gamma$ & 0.99 & 0.99 & 0.99 & 0.99 & 0.99 & 0.99 \\
		\midrule
        Input learning rate & 0.01 & 0.01 & 0.01 & 0.01 & 0.01 & 0.001 \\
        Circuit learning rate & 0.001 & 0.001 & 0.001 & 0.001 & 0.001 & 0.005 \\
        Output learning rate & 0.01 & 0.01 & 0.01 & 0.01 & 0.01 & 0.1 \\
		\midrule
        Classical learning rate & 0.001 & 0.001 & 0.001 & 0.001 & 0.001 & 0.01 \\
		\bottomrule
	\end{tabular}
\end{table}

\begin{table}[ht]
	\centering
	\caption{
		Statistics on the experiments executed in the small $3{\times}3$ environment for
		all three architectures and their configurations. The best statistical values (mean, median,
		minimum, maximum, and standard deviation) in terms of number of training steps for each
		architecture are marked bold, the best overall configuration for each architecture is marked
		with a green background.
	}
	\label{tab:res3}
	\begin{tabular}{cccccccc}
		\toprule
        \multicolumn{3}{c}{\underline{\textbf{NN}}}
		& \multicolumn{5}{c}{\textbf{No.\ of training steps}} \\
		\cmidrule(lr{.5em}){4-8}
		\textbf{Units} & $\boldsymbol{|\theta_{NN}|}$ & \textbf{Solved} & \textbf{Mean} &
		\textbf{Median} & \textbf{Min} & \textbf{Max} & \textbf{Std.} \\
		\cmidrule(lr{.5em}){1-1}
		\cmidrule(lr{.5em}){2-2} \cmidrule(lr{.5em}){3-3} \cmidrule(lr{.5em}){4-4}
		\cmidrule(lr{.5em}){5-5} \cmidrule(lr{.5em}){6-6} \cmidrule(lr{.5em}){7-7}
		\cmidrule(lr{.5em}){8-8}
        (8;8)     &    131 & 17/20 & 29,390 & 26,800 & 18,200 & 50,000 &  9,558 \\
        (16;8)    &    227 & 20/20 & 21,075 & 20,500 & 13,700 & 33,200 &  4,921 \\
        (16;16)   &    387 & 20/20 & 16,405 & 16,800 & 10,500 & 21,900 &  3,049 \\
        (32;16)   &    707 & 20/20 & 14,620 & 14,650 &  8,900 & 20,000 &  2,853 \\
        (32;32)   &  1,283 & 20/20 & 12,625 & 12,400 &  8,700 & 17,000 &  2,143 \\
        (64;32)   &  2,435 & 20/20 & 10,635 & 10,250 &  7,000 & 14,100 &  2,231 \\
        (64;64)   &  4,611 & 19/20 & 12,790 & 10,100 &  6,700 & 50,000 & 10,200 \\
        (128;64)  &  8,963 & 20/20 & 11,535 &  9,700 &  6,800 & 47,600 &  8,414 \\
        (128;128) & 17,411 & 20/20 &  8,885 &  8,650 &  5,800 & 11,900 &  1,746 \\
        \rowcolor{lightgreen}
        (256;128) & 34,307 & 20/20 &
        \textbf{7,495} &  \textbf{7,550} &  \textbf{5,200} &  \textbf{9,800} &  \textbf{1,324} \\
		\midrule
        \multicolumn{3}{c}{\underline{\textbf{PQC-1}}}
		& \multicolumn{5}{c}{\textbf{No.\ of training steps}} \\
		\cmidrule(lr{.5em}){4-8}
		\textbf{Layers} & $\boldsymbol{|\theta_{PQC}|}$ & \textbf{Solved} & \textbf{Mean} &
		\textbf{Median} & \textbf{Min} & \textbf{Max} & \textbf{Std.} \\
		\cmidrule(lr{.5em}){1-1}
		\cmidrule(lr{.5em}){2-2} \cmidrule(lr{.5em}){3-3} \cmidrule(lr{.5em}){4-4}
		\cmidrule(lr{.5em}){5-5} \cmidrule(lr{.5em}){6-6} \cmidrule(lr{.5em}){7-7}
		\cmidrule(lr{.5em}){8-8}
        12 & 156 & 20/20 & 24,670 & 25,200 & 13,400 & 31,800 &  4,034 \\
        15 & 192 & 20/20 & 18,620 & 18,800 &  9,000 & 24,000 &  3,402 \\
        18 & 228 & 20/20 & 19,345 & 19,250 & 14,900 & 24,000 &  \textbf{2,308} \\
        21 & 264 & 20/20 & 15,905 & 16,900 &  5,700 & 22,600 &  4,600 \\
        24 & 300 & 20/20 & 15,350 & 14,750 & 10,000 & 20,700 &  3,136 \\
        27 & 336 & 20/20 & 11,365 & 11,800 &  4,200 & 17,000 &  3,692 \\
        30 & 372 & 20/20 & 10,245 & 11,100 &  4,100 & 15,800 &  3,383 \\
        33 & 408 & 20/20 & 10,220 & 10,450 & \textbf{1,700} & 20,200 &  4,427 \\
        \rowcolor{lightgreen}
        36 & 444 & 20/20 & \textbf{9,150} & \textbf{9,350} &  2,000 & 16,500 &  3,746 \\
        39 & 480 & 20/20 & 10,060 & 10,050 &  4,800 & 13,900 & 2,905 \\
		\midrule
        \multicolumn{3}{c}{\underline{\textbf{PQC-3}}}
		& \multicolumn{5}{c}{\textbf{No.\ of training steps}} \\
		\cmidrule(lr{.5em}){4-8}
		\textbf{Layers} & $\boldsymbol{|\theta_{PQC}|}$ & \textbf{Solved} & \textbf{Mean} &
		\textbf{Median} & \textbf{Min} & \textbf{Max} & \textbf{Std.} \\
		\cmidrule(lr{.5em}){1-1}
		\cmidrule(lr{.5em}){2-2} \cmidrule(lr{.5em}){3-3} \cmidrule(lr{.5em}){4-4}
		\cmidrule(lr{.5em}){5-5} \cmidrule(lr{.5em}){6-6} \cmidrule(lr{.5em}){7-7}
		\cmidrule(lr{.5em}){8-8}
        8  & 156 & 20/20 & 20,270 & 20,000 & 14,300 & 26,300 &  3,608 \\
        10 & 192 & 20/20 & 14,930 & 16,750 &  3,900 & 22,900 &  5,231 \\
        12 & 228 & 20/20 & 16,530 & 15,850 &  7,100 & 25,000 &  4,155 \\
        14 & 264 & 20/20 & 10,995 & 11,750 &  2,600 & 22,100 &  5,188 \\
        16 & 300 & 20/20 &  9,300 &  8,750 &  1,200 & 20,600 &  4,175 \\
        18 & 336 & 20/20 & 10,385 &  9,450 &  2,600 & 19,300 &  4,553 \\
        20 & 372 & 20/20 & 10,390 & 10,400 &  1,800 & 18,000 &  4,164 \\
        22 & 408 & 20/20 &  9,490 &  8,400 &  2,300 & 20,300 &  5,181 \\
        \rowcolor{lightgreen}
        24 & 444 & 20/20 &
        \textbf{6,850} &  \textbf{6,650} &  \textbf{1,700} & \textbf{12,800} &  \textbf{3,146} \\
        26 & 480 & 20/20 &  9,515 &  8,800 &  4,400 & 15,300 &  3,347 \\
		\bottomrule
	\end{tabular}
\end{table}

\begin{table}[ht]
	\centering
	\caption{
		Statistics on the experiments executed in the medium sized $4{\times}4$ environment
		for all three architectures and their configurations. The best statistical values (mean,
		median, minimum, maximum, and standard deviation) in terms of number of training steps for
		each architecture are marked bold, the best overall configuration for each architecture is
        marked with a green background.}
	\label{tab:res4}
	\begin{tabular}{cccccccc}
		\toprule
		\multicolumn{3}{c}{\underline{\textbf{NN}}}
		& \multicolumn{5}{c}{\textbf{No.\ of training steps}} \\
		\cmidrule(lr{.5em}){4-8}
		\textbf{Units} & $\boldsymbol{|\theta_{NN}|}$ & \textbf{Solved} & \textbf{Mean} &
		\textbf{Median} & \textbf{Min} & \textbf{Max} & \textbf{Std.} \\
		\cmidrule(lr{.5em}){1-1}
		\cmidrule(lr{.5em}){2-2} \cmidrule(lr{.5em}){3-3} \cmidrule(lr{.5em}){4-4}
		\cmidrule(lr{.5em}){5-5} \cmidrule(lr{.5em}){6-6} \cmidrule(lr{.5em}){7-7}
		\cmidrule(lr{.5em}){8-8}
        (8;8)     &    131 & 16/20 & 37,315 & 37,050 & 19,600 & 50,000 &  9,329 \\
        (16;8)    &    227 & 17/20 & 35,910 & 34,650 & 23,700 & 50,000 &  7,477 \\
        (16;16)   &    387 & 17/20 & 24,565 & 23,350 &    900 & 50,000 & 13,387 \\
        (32;16)   &    707 & 17/20 & 28,060 & 25,100 &  7,200 & 50,000 & 11,359 \\
        (32;32)   &  1,283 & 16/20 & 24,315 & 19,800 &  7,200 & 50,000 & 13,548 \\
        (64;32)   &  2,435 & 20/20 & 15,055 & 16,550 &  6,300 & 28,300 &  5,543 \\
        (64;64)   &  4,611 & 20/20 & 14,740 & 15,050 &  6,900 & 22,200 &  4,205 \\
        (128;64)  &  8,963 & 19/20 & 18,370 & 15,000 &  5,400 & 50,000 & 11,092 \\
        (128;128) & 17,411 & 20/20 & 12,270 & \textbf{11,900} &  5,700 & 20,000 &  3,930 \\
        \rowcolor{lightgreen}
        (256;128) & 34,307 & 20/20 & \textbf{11,480} & 12,000 &
                \textbf{4,400} & \textbf{18,900} &  \textbf{3,800} \\
		\midrule
		\multicolumn{3}{c}{\underline{\textbf{PQC-1}}}
		& \multicolumn{5}{c}{\textbf{No.\ of training steps}} \\
		\cmidrule(lr{.5em}){4-8}
		\textbf{Layers} & $\boldsymbol{|\theta_{PQC}|}$ & \textbf{Solved} & \textbf{Mean} &
		\textbf{Median} & \textbf{Min} & \textbf{Max} & \textbf{Std.} \\
		\cmidrule(lr{.5em}){1-1}
		\cmidrule(lr{.5em}){2-2} \cmidrule(lr{.5em}){3-3} \cmidrule(lr{.5em}){4-4}
		\cmidrule(lr{.5em}){5-5} \cmidrule(lr{.5em}){6-6} \cmidrule(lr{.5em}){7-7}
		\cmidrule(lr{.5em}){8-8}
        12 & 156 & 18/20 & 41,395 & 42,400 & 28,700 & 50,000 &  5,260 \\
        15 & 192 & 20/20 & 29,905 & 31,700 &  9,300 & 46,100 &  9,254 \\
        18 & 228 & 20/20 & 27,055 & 30,950 &  2,400 & 39,500 & 11,105 \\
        21 & 264 & 20/20 & 24,815 & 26,750 &  4,000 & 36,000 &  8,303 \\
        24 & 300 & 20/20 & 27,795 & 27,250 & 22,200 & 36,000 &  \textbf{3,913} \\
        27 & 336 & 20/20 & 22,355 & 24,400 & 11,800 & 31,700 &  6,131 \\
        30 & 372 & 20/20 & 20,560 & 21,600 &  4,800 & 30,700 &  6,542 \\
        33 & 408 & 20/20 & 20,135 & 20,750 &  4,400 & 29,900 &  5,640 \\
        36 & 444 & 20/20 & 17,705 & 18,800 &  7,000 & \textbf{26,000} &  4,678 \\
        \rowcolor{lightgreen}
        39 & 480 & 20/20 & \textbf{16,880} & \textbf{18,600} &  \textbf{3,400} & 29,300 &  7,288 \\
		\midrule
		\multicolumn{3}{c}{\underline{\textbf{PQC-3}}}
		& \multicolumn{5}{c}{\textbf{No.\ of training steps}} \\
		\cmidrule(lr{.5em}){4-8}
		\textbf{Layers} & $\boldsymbol{|\theta_{PQC}|}$ & \textbf{Solved} & \textbf{Mean} &
		\textbf{Median} & \textbf{Min} & \textbf{Max} & \textbf{Std.} \\
		\cmidrule(lr{.5em}){1-1}
		\cmidrule(lr{.5em}){2-2} \cmidrule(lr{.5em}){3-3} \cmidrule(lr{.5em}){4-4}
		\cmidrule(lr{.5em}){5-5} \cmidrule(lr{.5em}){6-6} \cmidrule(lr{.5em}){7-7}
		\cmidrule(lr{.5em}){8-8}
        8  & 156 & 20/20 & 30,155 & 32,900 &  8,100 & 41,300 &  9,654 \\
        10 & 192 & 20/20 & 25,990 & 28,700 &  8,800 & 39,900 &  8,469 \\
        12 & 228 & 20/20 & 21,525 & 23,800 &  3,500 & 31,900 &  8,074 \\
        14 & 264 & 20/20 & 19,695 & 20,750 &  2,700 & 29,900 &  7,103 \\
        16 & 300 & 20/20 & 20,170 & 21,950 &  4,300 & 28,300 &  5,984 \\
        18 & 336 & 20/20 & 19,015 & 20,950 &  4,400 & 26,700 &  5,891 \\
        20 & 372 & 20/20 & 18,795 & 19,700 &  6,000 & 24,500 &  \textbf{4,346} \\
        22 & 408 & 20/20 & 17,065 & 17,150 &  5,500 & 26,600 &  5,652 \\
        \rowcolor{lightgreen}
        24 & 444 & 20/20 & \textbf{14,665} & 15,350 &  \textbf{2,000} & \textbf{25,000} &  6,068 \\
        26 & 480 & 20/20 & 15,875 & \textbf{15,050} &  7,200 & 25,200 &  5,164 \\
		\bottomrule
	\end{tabular}
\end{table}

\begin{table}[ht]
	\centering
	\caption{
        Statistics on the experiments executed in the large $5{\times}5$ environment for all three
        architectures and their configurations. The best statistical values (mean, median, minimum,
        maximum, and standard deviation) in terms of number of training steps for each architecture
        are marked bold, the best overall configuration for each architecture is marked with a green
        background. Configurations for which fewer than 15 runs succeeded are considered insufficient
        and are marked with a red background.}
	\label{tab:res5}
	\begin{tabular}{cccccccc}
		\toprule
		\multicolumn{3}{c}{\underline{\textbf{NN}}}
		& \multicolumn{5}{c}{\textbf{No.\ of training steps}} \\
		\cmidrule(lr{.5em}){4-8}
		\textbf{Units} & $\boldsymbol{|\theta_{NN}|}$ & \textbf{Solved} & \textbf{Mean} &
		\textbf{Median} & \textbf{Min} & \textbf{Max} & \textbf{Std.} \\
		\cmidrule(lr{.5em}){1-1}
		\cmidrule(lr{.5em}){2-2} \cmidrule(lr{.5em}){3-3} \cmidrule(lr{.5em}){4-4}
		\cmidrule(lr{.5em}){5-5} \cmidrule(lr{.5em}){6-6} \cmidrule(lr{.5em}){7-7}
		\cmidrule(lr{.5em}){8-8}
        \rowcolor{lightred}
        (8;8)     &    131 & 1/20 & 49,825 & 50,000 & 46,500 & 50,000 &    762 \\
        \rowcolor{lightred}
        (16;8)    &    227 & 4/20 & 49,515 & 50,000 & 42,500 & 50,000 &  1,641 \\
        \rowcolor{lightred}
        (16;16)   &    387 & 8/20 & 46,570 & 50,000 & 32,400 & 50,000 &  5,104 \\
        (32;16)   &    707 & 15/20 & 42,040 & 41,350 & 29,700 & 50,000 &  6,291 \\
        (32;32)   &  1,283 & 17/20 & 40,055 & 40,900 & 23,300 & 50,000 &  8,465 \\
        (64;32)   &  2,435 & 18/20 & 32,135 & 30,850 & 22,000 & 50,000 &  7,845 \\
        (64;64)   &  4,611 & 20/20 & 29,475 & 28,450 & 23,300 & 40,400 &  4,218 \\
        (128;64)  &  8,963 & 20/20 & 25,540 & 24,950 & 19,500 & \textbf{31,600} &  \textbf{3,132} \\
        (128;128) & 17,411 & 20/20 & 22,870 & 22,100 & \textbf{15,900} & 35,900 &  4,301 \\
        \rowcolor{lightgreen}
        (256;128) & 34,307 & 20/20 & \textbf{22,220} & \textbf{21,100} & 17,700 & 33,500 &  4,017 \\
		\midrule
		\multicolumn{3}{c}{\underline{\textbf{PQC-1}}}
		& \multicolumn{5}{c}{\textbf{No.\ of training steps}} \\
		\cmidrule(lr{.5em}){4-8}
		\textbf{Layers} & $\boldsymbol{|\theta_{PQC}|}$ & \textbf{Solved} & \textbf{Mean} &
		\textbf{Median} & \textbf{Min} & \textbf{Max} & \textbf{Std.} \\
		\cmidrule(lr{.5em}){1-1}
		\cmidrule(lr{.5em}){2-2} \cmidrule(lr{.5em}){3-3} \cmidrule(lr{.5em}){4-4}
		\cmidrule(lr{.5em}){5-5} \cmidrule(lr{.5em}){6-6} \cmidrule(lr{.5em}){7-7}
		\cmidrule(lr{.5em}){8-8}
        \rowcolor{lightred}
        12 & 156 & 4/20 & 48,005 & 50,000 & 31,400 & 50,000 &  4,644 \\
        \rowcolor{lightred}
        15 & 192 & 12/20 & 46,390 & 47,700 & 36,500 & 50,000 &  4,021 \\
        18 & 228 & 18/20 & 40,410 & 39,850 & 25,900 & 50,000 &  6,182 \\
        21 & 264 & 20/20 & 37,375 & 34,500 & 29,400 & 49,700 &  6,546 \\
        24 & 300 & 20/20 & 35,190 & 33,600 & 27,400 & 47,200 &  \textbf{5,216} \\
        27 & 336 & 19/20 & 35,685 & 32,900 & 19,300 & 50,000 &  7,881 \\
        30 & 372 & 20/20 & 30,995 & 31,400 &  8,500 & 46,600 &  9,123 \\
        33 & 408 & 20/20 & 28,155 & 27,900 & 19,100 & 38,500 &  5,752 \\
        36 & 444 & 20/20 & 29,260 & 29,300 & 15,800 & 47,000 &  6,614 \\
        \rowcolor{lightgreen}
        39 & 480 & 20/20 & \textbf{25,110} & \textbf{26,550} & \textbf{12,900} &
            \textbf{33,500} &  5,309 \\
		\midrule
		\multicolumn{3}{c}{\underline{\textbf{PQC-3}}}
		& \multicolumn{5}{c}{\textbf{No.\ of training steps}} \\
		\cmidrule(lr{.5em}){4-8}
		\textbf{Layers} & $\boldsymbol{|\theta_{PQC}|}$ & \textbf{Solved} & \textbf{Mean} &
		\textbf{Median} & \textbf{Min} & \textbf{Max} & \textbf{Std.} \\
		\cmidrule(lr{.5em}){1-1}
		\cmidrule(lr{.5em}){2-2} \cmidrule(lr{.5em}){3-3} \cmidrule(lr{.5em}){4-4}
		\cmidrule(lr{.5em}){5-5} \cmidrule(lr{.5em}){6-6} \cmidrule(lr{.5em}){7-7}
		\cmidrule(lr{.5em}){8-8}
		\rowcolor{lightred}
        8  & 156 & 13/20 & 43,925 & 45,700 & 32,100 & 50,000 &  6,303 \\
        10 & 192 & 18/20 & 41,645 & 42,500 & 28,800 & 50,000 &  6,744 \\
        12 & 228 & 20/20 & 35,675 & 35,000 & 27,100 & 47,600 &  6,372 \\
        14 & 264 & 20/20 & 33,885 & 33,400 & 22,400 & 45,200 &  5,778 \\
        16 & 300 & 20/20 & 32,625 & 33,700 & 23,000 & 43,300 &  5,588 \\
        18 & 336 & 20/20 & 30,765 & 30,300 & 22,600 & 42,200 &  5,212 \\
        20 & 372 & 20/20 & 29,895 & 31,250 & 20,500 & 39,300 &  \textbf{5,037} \\
        22 & 408 & 20/20 & 24,530 & 24,900 & 12,800 & \textbf{38,600} &  5,762 \\
        24 & 444 & 20/20 & 27,575 & 28,650 &  \textbf{9,400} & 38,800 &  7,334 \\
        \rowcolor{lightgreen}
        26 & 480 & 20/20 & \textbf{23,635} & \textbf{21,800} &  9,700 & 41,600 &  7,535 \\
		\bottomrule
	\end{tabular}
\end{table}

\clearpage
\twocolumn

\begin{IEEEbiography}[{%
    \includegraphics[width=1in,height=1.25in,clip,keepaspectratio]{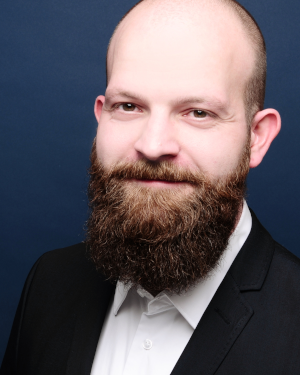}
}]{Hans Hohenfeld}
received his B.\,Sc.\ and M.\,Sc.\ degrees in computer science from the \mbox{FernUniversität} in
Hagen in 2014 and 2017. Before joining the Robotics Research Group at University of
Bremen in 2017, he worked as software engineer since 2008. Currently
he is pursuing a Ph.\,D.\ degree at the University of Bremen. His research interests include
hardware efficient quantum algorithms for machine learning as well as quantum information and
quantum algorithms in learning and decision problems for autonomous agents.
\end{IEEEbiography}

\begin{IEEEbiography}[{%
    \includegraphics[width=1in,height=1.25in,clip,keepaspectratio]{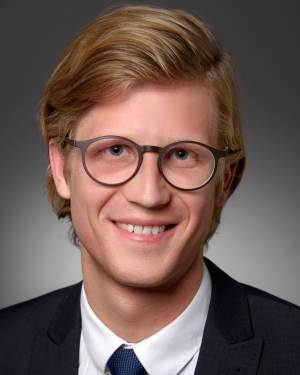}
}]{Dirk Heimann}
received his B.\,Sc.\ and M.\,Sc.\ degrees in physics from the Leibniz Universität Hannover
in 2014 and 2017, respectively. Before joining the Robotics Research Group at University of Bremen
in 2020, he worked as software consultant in the industry. He is currently pursuing a Ph.\,D.\
degree at the University of Bremen. His research interests include quantum algorithms for machine
learning as well as quantum optimal control.
\end{IEEEbiography}

\begin{IEEEbiography}[{%
    \includegraphics[width=1in,height=1.25in,clip,keepaspectratio]{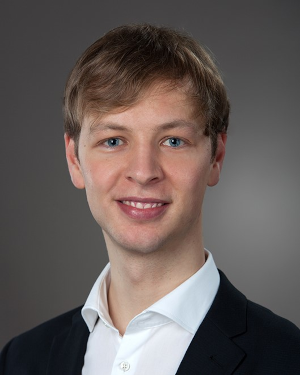}
}]{Felix Wiebe}
received his B.\,Sc.\ and M.\,Sc.\ degrees in physics from the \mbox{Georg August University} in
Göttingen in 2014 and 2018, respectively. In 2018, he joined the Mechanics and Control team of the
German Research Center for Artificial Intelligence (DFKI) in the Robotics Innovation Center (RIC) in
Bremen. Since 2020, he is pursuing a Ph.\,D.\ in the field of optimal control for underactuated
robots. His research interests include robotics, optimal control, model predictive control,
optimization algorithms, and quantum optimal control.
\end{IEEEbiography}

\begin{IEEEbiography}[{%
    \includegraphics[width=1in,height=1.25in,clip,keepaspectratio]{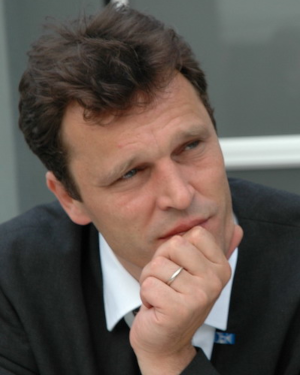}
}]{Frank Kirchner}
received the degree in computer science and neurobiology and the Dr.\,rer.\,nat.\ degree in computer
science from the University of Bonn, in 1994 and 1999, respectively. Since 1994, he has been a
Senior Scientist with the Gesellschaft für Mathematik und Datenverarbeitung (GMD), Sankt Augustin.
He has been a Senior Scientist with the Department for Electrical Engineering, Northeastern
University, Boston, MA, USA, since 1998, where he was appointed as an Adjunct and then a Tenure
Track Assistant Professor, in 1999. Since 2002, he has been a Full Professor with the University of
Bremen. Since December 2005, he has been the Director of the Robotics Innovation Center (RIC),
Bremen.
\end{IEEEbiography}
\EOD
\end{document}